\documentclass[times, review, 10pt]{elsarticle}
\usepackage{amssymb}
\usepackage{amsmath}
\usepackage{textcase}
\usepackage{subcaption}
\usepackage{graphicx}
\usepackage[colorlinks=true, allcolors=blue]{hyperref}
\usepackage{comment}
\usepackage{multirow}
\usepackage{booktabs}
\usepackage{xcolor}
\usepackage{float}
\usepackage[font={footnotesize}]{caption}
\usepackage[colorinlistoftodos]{todonotes}
\journal{Biomedical Signal Processing and Control}
\begin{document}

\begin{frontmatter}
\title{Fusion Techniques of Time Frequency-based Images to Predict the Outcome of rTMS Depression Therapy}
\author{Wael Korani\corref{cor1}\fnref{label1}}
\ead{wael.korani@unt.edu}
\cortext[cor1]{Corresponding author}
\affiliation[label1]{organization={Department of Data Science, University of North Texas},
state={Texas},
country={USA}}
\author{Md Fahimul Kabir Chowdhury\fnref{label2}}
\ead{MdFahimulKabirChowdhury@my.unt.edu}
\affiliation[label2]{organization={Department of Computer Science, University of North Texas},
state={Texas},
country={USA}}
\author{Mohammed Aledhari\fnref{label1}}
\ead{Mohammed.Aledhari@unt.edu}
\author{Reza Rostami \fnref{label3}}
\ead{reza.rostami@ut.ac.ir}
\affiliation[label3]{organization={Department of Psychiatry, University of Tehran}, 
            city={Tehran},
            country={Iran}}
\author{Reza Kazemi \fnref{label3}}
\ead{reza.kazemi@ut.ac.ir}
\begin{abstract}
Depression is a mental condition that can lead to suicide and self-harm. Predicting the outcome of depression treatment is one of the most difficult tasks for clinicians. Among various treatment options, repetitive Transcranial Magnetic Stimulation (rTMS) is a widely used non-invasive method. Predicting rTMS response using Electroencephalogram (EEG) data is difficult because of high inter-subject variability and limited features from single-domain analysis. We introduce two fusion techniques, montage and blending, to overcome these limitations and extract richer features from EEG-derived Time-Frequency (TF) images. We then propose a lightweight custom Convolutional Neural Network (CNN) trained on fused TF representations.
\textcolor{black}{We use a primary dataset of 15 patients and a secondary dataset of 46 patients. We run two sets of experiments. The first set uses segment-level 10-fold cross-validation. In this setup segments from the same patient can appear in both training and testing. The Montage CWT\_ST fusion reaches 99.90\% accuracy on the primary dataset and 91.90\% on the secondary dataset. The second set uses strict subject-disjoint cross-validation. All segments of a patient stay in one fold and no patient appears in both training and testing. Performance collapses. We test four time-frequency methods, six fusion mechanisms, and fourteen model architectures. With one exception, every configuration on both cohorts falls between AUC 0.31 and 0.54 and every 95\% confidence interval contains 0.5. A patient-level permutation test on the best standalone method returns $p = 0.703$. The best subject-level result is Montage CWT\_ST on the primary cohort, which reaches AUC $0.874 \pm 0.183$ and 82.7\% accuracy. The same fusion strategy on the same pair of methods reaches 0.514 on the secondary cohort, so the mechanism behind this difference is not yet established and we identify it as the main target for follow-up work. The gap between the two experimental setups is the main result. Segment-level validation raises accuracy from chance to near-perfect on this data. Similar figures in the EEG literature should therefore be read together with the fold construction used to produce them.}
\end{abstract}
\begin{graphicalabstract}
\begin{figure}[ht]
\centering
\includegraphics[width=1\textwidth]{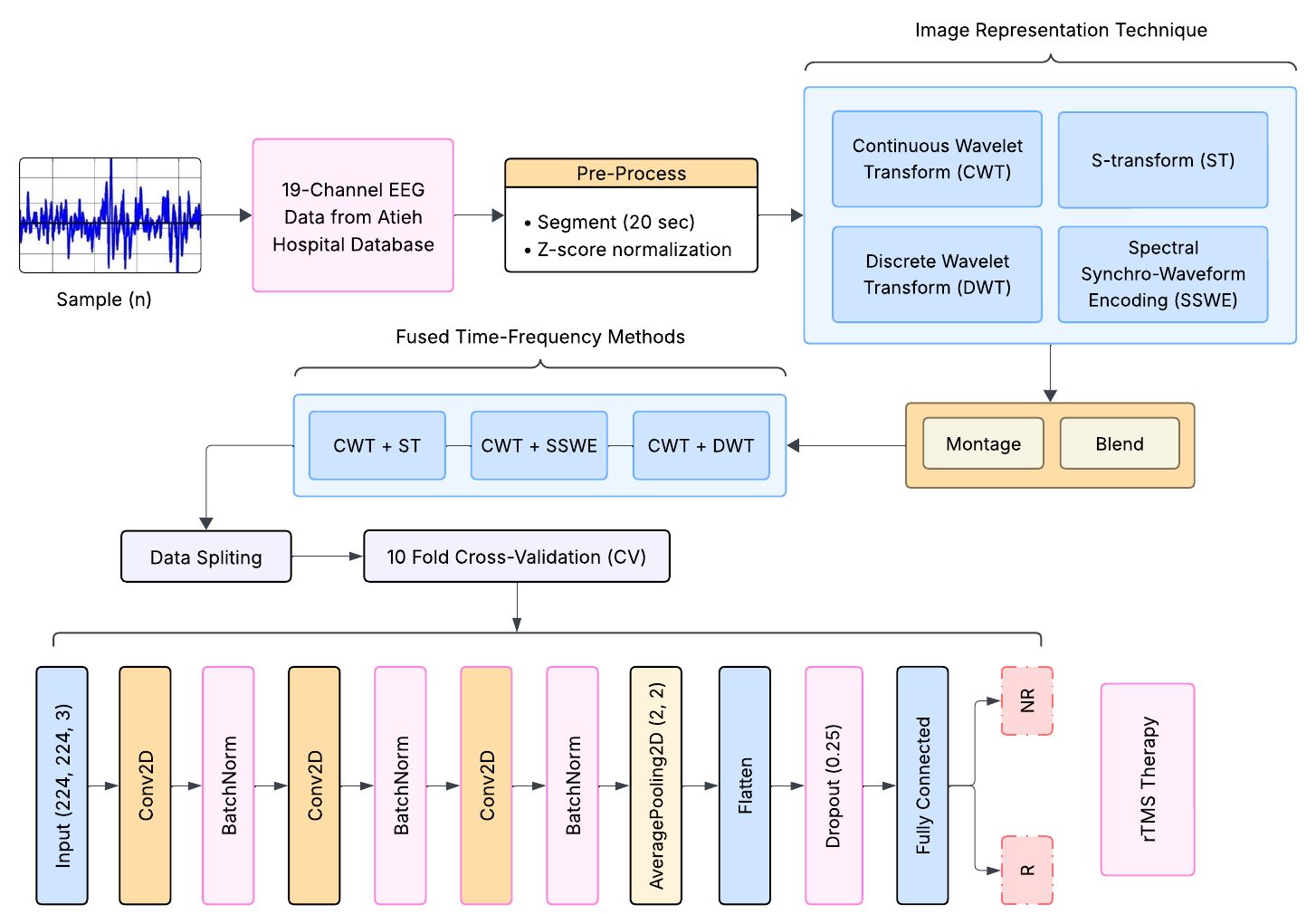}
\caption{A summary of the EEG data preprocessing workflow and the transformation of signals into image representations using multiple time-frequency techniques, which are then used as inputs for the proposed custom CNN model}
\label{blockD}
\end{figure}
\end{graphicalabstract}
\begin{highlights}
\item Proposing two fusion techniques: blend and montage to extract more features of EEG signals
\item Developing an accurate and robust lightweight custom CNN model to predict the outcomes of depression therapies
\item Generating fused time frequency images to predict the outcome of depression therapy
\item Comparing the accuracy of our model using fused time frequency techniques against traditional time frequency methods
\end{highlights}
\end{frontmatter}
\section{Introduction}
Major Depressive Disorder (MDD) is a serious mental health condition that affects people of all ages, genders, and backgrounds. It is one of the leading causes of disability worldwide \cite{martino2017connection}. The cause of MDD is difficult to identify because it results from a mix of genetic, biological, environmental, and psychological factors \cite{marx2023major}. Researchers find that changes in key neurotransmitters such as serotonin, dopamine, and norepinephrine, as well as problems with the body's stress response system (the HPA axis), play a role in depression \cite{correia2024advancements}. Structural changes in brain areas that regulate mood, such as the prefrontal cortex and hippocampus, are also linked to the condition \cite{drevets2008brain}. Depression shows up in many ways. People with depression feel persistently sad, empty, or hopeless, and lose interest in things they once enjoyed, which is called anhedonia \cite{watson2020understanding}. They may also notice changes in appetite, weight, or sleep. In severe cases, depression brings thoughts of death or suicide \cite{obuobi2021prevalence}.
Despite the heavy impact of MDD, effective treatments are available. Drug treatments such as Selective Serotonin Reuptake Inhibitors (SSRIs), serotonin-norepinephrine reuptake inhibitors (SNRIs), and tricyclic antidepressants (TCAs) are commonly prescribed. For patients who do not respond to these, repetitive Transcranial Magnetic Stimulation (rTMS), a non-invasive brain stimulation technique, is gaining more attention as a safe alternative.
rTMS is a non-invasive technique that uses magnetic pulses to stimulate brain areas linked to mood regulation. It offers hope for patients with treatment-resistant depression. Several studies confirm that rTMS has limited side effects and good overall effectiveness. To prescribe rTMS, doctors conduct a careful assessment and a series of tests to improve diagnostic accuracy.
Selecting the best course of depression therapy remains a trial and error process. This long-standing clinical challenge drives an increased focus on treatment response prediction using different types of data, including electroencephalogram (EEG) signals and machine learning (ML) \cite{chekroud2021promise}. ML including deep learning (DL) has become a powerful tool in mental health treatment. These techniques use algorithms and data to improve everything from diagnosis to treatment personalization.
Deep learning uses complex neural networks to analyze data such as functional MRI scans and EEG signals. This helps researchers understand the brain activity linked to depression and how patients respond to treatment.
Feature fusion combines valuable information from different sources into a single representation, which is more useful for classification tasks. Fusion can be done in several ways: fusing different EEG features, time-domain features like zero-crossing rate, different time-frequency methods, different frequency bands (alpha, beta, delta, theta, gamma), or wavelet coefficients \cite{alyasseri2021eeg}. Effective strategies for integrating these features include concatenation, weighted averaging, and attention mechanisms. Recent studies show that fusing features improves model classification accuracy by extracting richer and more robust representations \cite{zhang2024scheme}. DL and specifically CNN, with their strong ability to extract and reconstruct features, make fused images more effective. Fusion techniques provide a more complete view of neural activity and result in better classification accuracy compared to single features alone \cite{zhang2021image,chowdhury2025brain}.
{\color{black}
The development of reliable computer-aided diagnostic (CAD) systems for complex physiological conditions requires descriptive feature representations and rigorous validation strategies. Recent literature shows that advanced machine learning architectures, enhanced feature selection methods, and metaheuristic optimization successfully resolve the complex non-linear dynamics in heterogeneous biomedical datasets. These applications cover brain tumor MRI classification, Alzheimer's disease connectivity mapping, diabetic retinopathy tracking, and multi-stage growth detection \cite{ozkan2026optimization, ozccelik2023overcoming, cicek2025new, ozccelik2024machine}. This study follows this push toward methodological rigor and addresses the historical limitations of standalone time-frequency analysis in psychiatric classification models.
}
In this study, we predict how patients respond to rTMS depression therapy for MDD, and we identify responders (R) and non-responders (NR). We generate EEG-derived time-frequency images using different time-frequency methods. The generated images are used to propose a lightweight custom CNN. To improve classification accuracy, we fuse two different time-frequency analysis methods. We use four time-frequency methods: Continuous Wavelet Transform (CWT), Discrete Wavelet Transform (DWT), S-Transform (ST), and Wavelet Synchrosqueezing Transform (SSWE). We combine two of these methods in two ways: blending and montage. These datasets then train our proposed custom CNN model. Our CNN offers both computational efficiency and a lightweight architecture, and it outperforms various ML, transfer learning (TL), and hybrid models. This research highlights the power of fusing EEG-derived time-frequency images to predict a patient's response to rTMS therapy. This approach sheds new light on treatment predictions for MDD patients and opens the way for more personalized treatment plans in psychiatric care.
This paper makes several important contributions:
{\color{black}
\begin{itemize}
\item We present the first subject-disjoint evaluation of EEG-based rTMS response prediction that we are aware of. Every study we cite reports k-fold accuracy without stating whether folds were built at the patient level or the segment level. This paper gives the first properly validated estimate of what this family of methods achieves when no patient appears in both training and testing.
\item We measure what fold construction is worth on clinical EEG. Eight architectures are run under both protocols with the same code and the same training settings, changing only how folds are built. All eight move from a patient-level AUC of exactly 1.000 to chance, across parameter counts spanning three orders of magnitude.
\item We generate TF images from EEG signals using four methods (CWT, DWT, ST, and SSWE) and introduce two image-level fusion strategies, blending and montage. We compare these against four further fusion mechanisms: channel-wise stacking, feature-level concatenation, attention-weighted fusion, and late decision fusion.
\item We compare our lightweight CNN against thirteen alternative architectures under one protocol. These span five pretrained vision models, three EEG-specific models (EEGNet, EEGNetV2, SleepEEGNet), a CNN-LSTM hybrid, and four classical machine learning classifiers.
\item We report bootstrap confidence intervals with the patient as the unit of analysis, Brier scores for calibration, and patient-level permutation testing. These are not standard in this literature and they are what allow the results to be interpreted at this sample size.
\end{itemize}
}
\section{Literature Review}
\subsection{Feature Fusion}
In \cite{chen2025mdd}, Chen et al. proposed a feature fusion technique and improved feature selection for diagnosing MDD. Seven functional connectivity matrices were extracted and combined into a single vector to create a fusion feature. Seven classifiers were tested for MDD diagnosis, and the proposed method outperformed existing techniques, achieving an accuracy of 88.73\%. In \cite{zhang2023spatial}, Zhang et al. developed a spatial-temporal EEG fusion framework using a neural network for the detection of MDD. The temporal EEG data was mapped to a spatial functional brain network using the phase lag index, and spatial features were extracted from this network using 2D convolutional neural networks. The spatial-temporal EEG features were then fused to enhance data diversity, which results in a performance of 96.33\%. In \cite{xing2024adaptive}, Xing et al. proposed EMO-GCN, a multimodal depression detection method based on an adaptive multi-graph neural network. EMO-GCN uses graph-based methods to model EEG and patient interview audios, enabling the extraction of features from both modalities and revealing their potential correlations. The model achieved performance on the MODMA dataset, reaching an accuracy of 96.30\%.
In \cite{cui2023multiview},  Cui et al. introduced a novel EEG-based framework for detecting MDD, named the multi-view sparse dynamic graph convolution-based region-attention feature fusion network (MV-SDGC-RAFFNet). The region-attention feature fusion network (RAFFNet) then applies different attention weights for brain regions. The model was validated on two public MDD datasets, MODMA and HUSM, achieved a classification accuracy of 95.53\% and 99.19\%. In \cite{duan2020machine}, Duan et al. proposed a novel approach for identifying MDD by fusing interhemispheric asymmetry and cross-correlation with EEG signals. The approach was tested on 32 subjects, including 16 MDD patients and 16 healthy controls (HCs). Structural and connectivity features were extracted from the $\theta$-, $\alpha$-, and $\beta$-frequency bands of the preprocessed EEG signals. These features were then fed into three classifiers to determine the best feature set for classification, with mixed features yielding an accuracy of 94.13\%.
\subsection{Machine Learning}
In \cite{bailey2018responders}, Bailey et al. conducted a study with 39 patients and 32 patients suffering from treatment-resistant depression, who received 15 sessions of unilateral left 10 Hz rTMS over 3 weeks. The goal was to predict treatment response, categorizing patients as responders versus non-responders. The study employed a linear Support Vector Machine (SVM) model, achieving an impressive accuracy of 91\% and  86.66\%. In this study \cite{ebrahimzadeh2023machine}, Ebrahimzadeh et al. a machine learning-based strategy was developed to predict the therapeutic outcome of rTMS treatment. Resting-state EEG data were recorded from 88 MDD patients before treatment, and after 7 weeks of rTMS, 46 patients were classified as responders. A Genetic Algorithm (GA) selected the most relevant features, and K-Nearest Neighbor (KNN), SVM, and Multilayer Perceptron (MLP) classifiers were used. The method achieved high performance, with SVM yielding an accuracy of 94.31\%.
In \cite{corlier2019changes}, Corlier et al. investigated 109 patients with MDD, who underwent 15 sessions of 10 Hz left DLPFC rTMS over 3 weeks. Elastic Net was used for classification, yielding accuracy between 61.8\% and 79.2\%. The area under the curve (AUC) ranged from 0.52 to 0.77, indicating moderate predictive ability. In \cite{zhaopredicting} this study, Zhao et al. aimed to improve the prediction of rTMS treatment outcomes in MDD by combining EEG and clinical features. A total of 74 MDD patients who responded to rTMS and 43 who did not were included. The study used eight baseline EEG metrics and clinical data to train seven machine learning models for classifying responders and non-responders. The best performance was achieved by combining phase locking value (PLV) and clinical features with a SVM, resulting in an accuracy of 97.33\%.
In \cite{hasanzadeh2019prediction}, Hasanzadeh et al. proposed a study with 46 patients diagnosed with MDD, who underwent 5 sessions of 10 Hz left DLPFC rTMS. The study aimed to identify responders versus non-responders, as well as differentiate between remission and non-remission. The classification was performed using the kNN algorithm, achieving an accuracy range of 76.1\% to 91.3\%. In \cite{cao2018identifying}, Cao et al. conducted a study with 37 patients suffering from treatment-resistant depression. The aim was to predict responders versus non-responders. Several classification models were used, including Linear Discriminant Analysis (LDA), Non-Metric Space Classifier (NMSC), kNN, SVM etc. The overall accuracy was 78.4\%, with the best performance achieved using the SVM model with a radial kernel.
In \cite{cook2020prospective}, Cook et al. conducted a study with 180 patients diagnosed with MDD, who participated in an 8-week trial of escitalopram (10 mg) or bupropion (150 mg). The trial included a 1-week single-blind phase with escitalopram, followed by 7 weeks of double-blind treatment. LDA was used for classification, achieving an accuracy of 64.4\%. In \cite{mumtaz2017wavelet}, Mumtaz et al. conducted an open-label trial with 34 patients diagnosed with MDD who were treated with an SSRI. The study aimed to predict responders versus non-responders. Logistic Regression (LR) was used for classification, achieving an accuracy of 87.5\%.
In \cite{zhdanov2020use}, Zhdanov et al. conducted a study with 122 patients diagnosed with MDD, who received 8 weeks of open-label escitalopram treatment (10--20 mg). The study aimed to predict responders versus non-responders. SVM with a radial kernel was used for classification. The model achieved an accuracy of 79.2\% using baseline EEG data, and improved to 82.4\% when both baseline and week 2 EEG data were used.
In \cite{erguzel2016machine}, Erguzel et al. conducted a study with 147 patients suffering from treatment-resistant depression. The aim was to predict responders and non-responders. The study used three classification models such as Artificial Neural Network (ANN), SVM, and Decision Tree (DT), achieving an accuracy range of 78.3\% to 86.4\%. The best performance was observed using the SVM model, with a balanced accuracy range of 54.71\% to 75.42\%.
\subsection{Convolutional Neural Network}
In \cite{kang2020deep}, Kang et al. approached a diagnosing depression using EEG data, focusing on the asymmetry feature, which is a key biomarker of depression. The study introduced a deep-asymmetry methodology that transforms the EEG asymmetry feature into a matrix image. This image was then used as input for a CNN. Which achieved an accuracy of 98.85\%. In \cite{loh2022decision}, Loh et al. proposed a deep CNN applied Short-Time Fourier Transform (STFT) to EEG signals to generate spectrogram images, which were then processed by the CNN for automated MDD detection. The dataset included 34 MDD patients and 30 healthy subjects from a public database. The model achieved classification accuracy of 99.58\%. In \cite{seal2021deprnet}, Seal et al. proposed a deep learning-based DeprNet for classifying EEG data from depressed and normal subjects to help diagnose depression. DeprNet was evaluated in two experiments: recordwise split and subjectwise split. Using recordwise, DeprNet achieved a remarkable accuracy of 99.37\%. However, with subjectwise split data, the model achieved an accuracy of 91.4\%, indicating potential overfitting with a smaller subject pool.
In \cite{li2019eeg} , Li et al. proposed a computer-aided detection (CAD) system using a convolutional neural network (ConvNet) for recognizing mild depression. To enhance its clinical applicability, transfer learning was employed. The proposed system achieved an accuracy of 85.62\% for classifying mild depression and normal controls using 24-fold cross-validation. In \cite{sandheep2019performance}, Sandheep et al. explored the use of CNNs, to classify depression. The CNN was trained with EEG data from 30 normal and 30 depressed individuals, achieving high classification accuracy. Specifically, the network attained 99.31\% accuracy when classifying depression from EEG signals recorded from the right hemisphere and 96.3\% from the left hemisphere, following ten-fold cross-validation. In \cite{acharya2018automated}, Acharya et al. presented a novel CNN, for EEG-based screening of depression. The model was tested with EEG data from 15 normal and 15 depressed patients, achieving accuracies of 93.5\% for left hemisphere data and 96.0\% for right hemisphere data.
In \cite{dang2020multilayer}, Dang et al. developed a frequency-dependent multilayer brain network combined with a deep CNN. A multilayer deep CNN architecture was designed to analyze the brain network topology features. Experimental results on a publicly available MDD dataset demonstrated that the proposed approach achieved an accuracy of 97.27\%. In \cite{shovon2019classification}, Shovon et al. introduced a Motor Imagery EEG signal classification framework based on CNN to improve classification accuracy for Brain-Computer Interface (BCI) systems. The CNN was designed to handle three-channel EEG signals achieving an average classification accuracy of 89.19\% on dataset 2b and 97.7\% accuracy for subject 7 on dataset III.
\subsection{Transfer Learning (TL)}
In \cite{shahabi2023predictiona}, Shahabi et al. used 19-channel EEG data from 34 patients. Effective Connectivity matrices were derived from four EEG frequency bands to create connectivity images. Five pre-trained CNNs were used to classify responders and non-responders, with VGG16 and DenseNet121 achieving the highest accuracy of 89.22\%. An ensemble model combining all CNNs with Differential Evolution Optimization (DEO) improved performance to 92.28\% accuracy. In \cite{shahabi2023convolutional}, shahabi et al. proposed a deep learning model to predict response to rTMS in MDD patients. They used pre-treatment EEG signals from the public TDBRAIN dataset and 46 proprietary MDD subjects, creating time-frequency representations with CWT. Among several classifiers, the best performance was achieved in the Frontal brain region with the TL-LSTM-Attention model based on EfficientNetB0, yielding 97.1\% accuracy.
\subsection{Artificial Neural Network (ANN)}
In \cite{erguzel2015neural}, Erguzel et al. conducted a study with 147 patients who had treatment-resistant depression, receiving 18 sessions of 25 Hz left PFC rTMS. The study aimed to identify responders (defined as a \{$\ge$\}50\% decrease in HAM-D scores after 3 weeks of treatment) versus non-responders. The classification was performed using a Backpropagation Neural Network (BPNN), achieving an accuracy of 89.12\%. Another study by \cite{erguzel2015feature}, Erguzel et al. shows with 55 patients diagnosed with MDD, who received 18 sessions of 25 Hz left PFC rTMS. The study focused on identifying responders (defined as a \{$\ge$\}50\% decrease in HAM-D scores after 3 weeks of treatment) versus non-responders. An Artificial Neural Network (ANN) was used for classification, achieving an accuracy of 89.09\%.
\subsection{Hybrid Architecture}
In \cite{adamson2022sex}, Adamson et al. approached a Convolutional Long Short-Term Memory (LSTM)-based deep learning architecture was used. The dataset consisted of 50 patients, and the focus was on predicting their response to rTMS treatment. The innovative approach involved generating five distinct classification models where the results showed high accuracy, with test accuracies of 93.3\%, 98\%, 95.2\%, 99.2\%, and 96.6\% for the respective models. In \cite{shahabi2023prediction}, Shahabi et al. conducted a hybrid model to predict the response to rTMS treatment using raw EEG signals. The model combined pre-trained CNNs, BLSTM cells, creating a powerful Transfer Learning (TL)-BLSTM hybrid. An ensemble of these models was formed using weighted majority voting. The ensemble model achieved 98.51\% accuracy.
\section{Materials and Methods}
Our experiments are performed on an NVIDIA H100 GPU system, which uses high-bandwidth memory and advanced tensor cores optimized for AI workloads. The powerful parallel processing of the H100 significantly accelerates both the time-frequency fusion analysis and the deep learning pipeline. An overview of the proposed framework is presented in Fig~\ref{blockD}.
\begin{figure}[ht]
\centering
\includegraphics[width=1\textwidth]{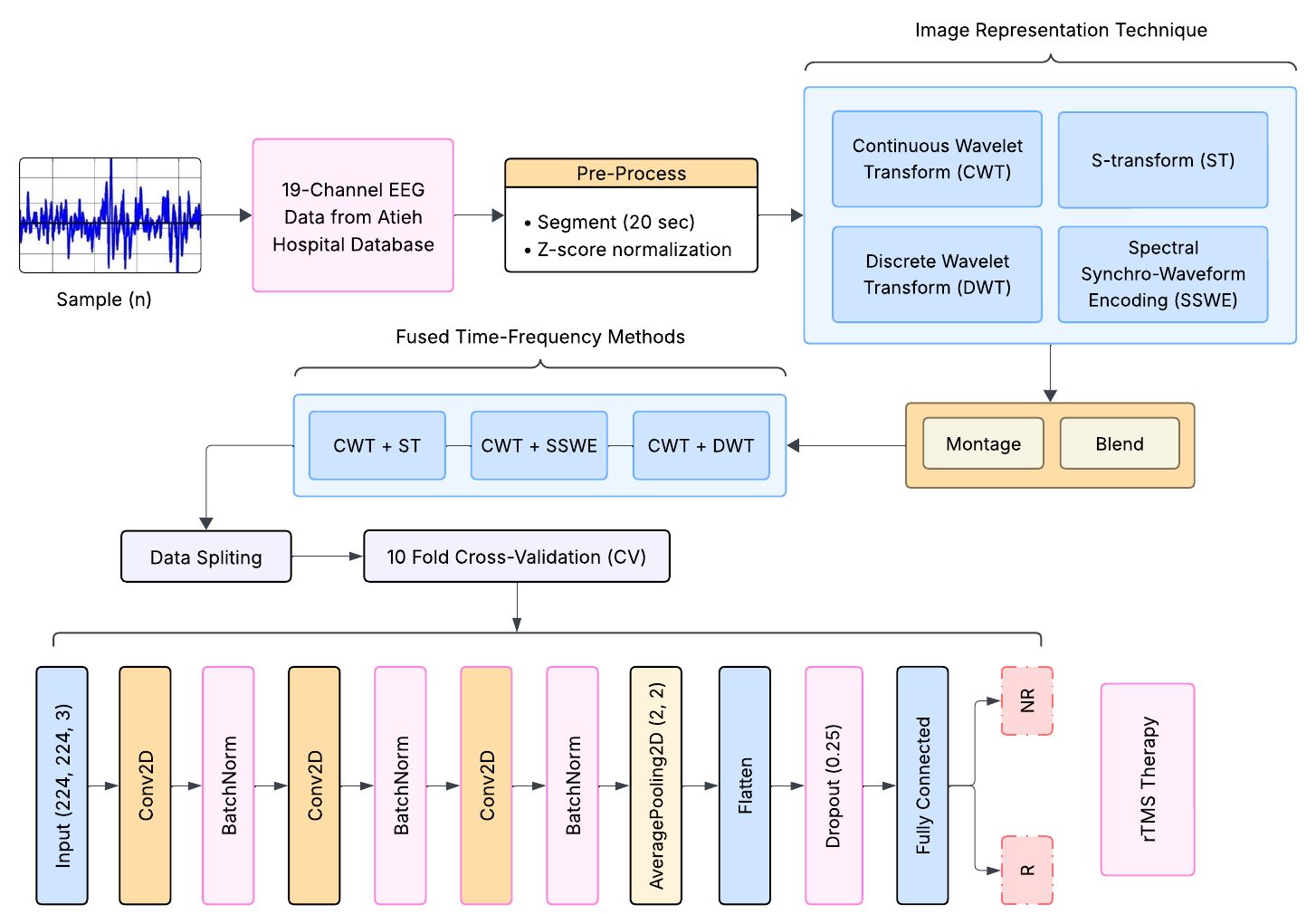}
\caption{A summary of the EEG data preprocessing workflow and the transformation of signals into image representations using multiple time-frequency techniques, which are then used as inputs for the proposed custom CNN model}
\label{blockD2}
\end{figure}
\subsection{Datasets and Clinical Characteristics}
{
\color{black}
We use a proprietary EEG database recorded at Atieh Hospital in Tehran, Iran. All clinical diagnoses of MDD are made by a certified psychiatrist and psychologist using standard diagnostic criteria. The database has two cohorts: a primary cohort of 15 patients and a secondary cohort of 46 patients. The two cohorts are recorded at different time periods and form independent datasets.

Patients are included if they are aged 18 to 60 years, have a confirmed MDD diagnosis based on the Structured Clinical Interview for DSM-5 (SCID-5-CV) conducted by a psychologist and psychiatrist, have a BDI-II score above 21, and are right-handed. Patients are excluded if they have substance abuse, suicidal thoughts, chronic medical conditions, a history of head injury with prolonged loss of consciousness, a history of neurosurgery, seizures or epilepsy in themselves or first-degree relatives, previous rTMS treatment, pregnancy or breastfeeding, or intracranial implants or metallic objects near the head. All patients stop their antidepressant medication before rTMS starts. No patient takes antidepressant medication during the treatment period, so rTMS is the only active treatment.

We measure clinical outcomes using the Beck Depression Inventory (BDI-II) at baseline (before rTMS) and after a four-week course of treatment. Patients with a $\geq$50\% reduction in BDI-II score after treatment are classified as responders (R). The rest are labeled non-responders (NR). A mental health specialist verifies all outcomes.
The primary dataset has 9 responders and 6 non-responders. BDI-II scores before and after treatment are shown in Table~\ref{tab:clinical_small}. Age, gender, and duration of depression are recorded for this pilot cohort, but these records are not available to us. The secondary dataset has 23 responders and 23 non-responders, with full demographic and clinical characteristics shown in Table~\ref{tab:clinical_large}.
\begin{table}[H]
\scriptsize
\centering
\begin{tabular}{|l|c|c|}
\hline
\textbf{Metric} & \textbf{Responders (R, $n=9$)} & \textbf{Non-Responders (NR, $n=6$)} \\
\hline
BDI-II Before rTMS (Mean $\pm$ SD) & 27.56 $\pm$ 8.19 & 33.83 $\pm$ 12.08 \\
BDI-II After rTMS (Mean $\pm$ SD)  & 7.44 $\pm$ 4.81  & 22.33 $\pm$ 10.59 \\
Mean Improvement (\%)             & 73.20 $\pm$ 13.57 & 33.06 $\pm$ 11.14 \\
\hline
\end{tabular}
\caption{Summary of BDI-II scores for the primary dataset ($N=15$).}
\label{tab:clinical_small}
\end{table}
\begin{table}[H]
\scriptsize
\centering
\begin{tabular}{|l|c|c|c|}
\hline
\textbf{Characteristic} & \textbf{Responders (R, $n=23$)} & \textbf{Non-Responders (NR, $n=23$)} & \textbf{$p$-Value} \\
\hline
Age (Mean $\pm$ SD) & 30.87 $\pm$ 12.00 & 39.00 $\pm$ 14.16 & 0.052 \\
Gender (Male/Female) & 8/15 & 8/15 & 0.900 \\
BDI Score Before Therapy & 32.5 $\pm$ 9.3 & 28.1 $\pm$ 9.4 & 0.080 \\
BDI Score After Therapy & 8.6 $\pm$ 5.9 & 23.1 $\pm$ 8.4 & $<0.001$ \\
Length of Depressive Episode (Years) & 6.5 $\pm$ 8.2 & 7.9 $\pm$ 7.8 & 0.270 \\
\hline
\end{tabular}
\caption{Demographic and clinical characteristics of the secondary dataset ($N=46$).}
\label{tab:clinical_large}
\end{table}
}
\subsection{rTMS Treatment Protocol and EEG Preprocessing}
{\color{black}
rTMS sessions are conducted five times per week over four consecutive weeks, totaling 20 sessions. Stimulation is delivered over the left dorsolateral prefrontal cortex (DLPFC) at 10 Hz and 110\% of the patient's resting motor threshold. Each session consists of 3,000 magnetic pulses.

Resting-state EEG is recorded before the first rTMS session using 19 electrodes placed according to the international 10-20 system, at a sampling rate of 500 Hz. Electrodes A1 and A2 are used as references (linked ears reference). The raw EEG signals are filtered using a Butterworth band-pass filter (0.5--70 Hz) and a notch filter at 50 Hz to remove baseline drift and power-line noise. Independent Component Analysis (ICA) is applied to remove eye and muscle artifacts. Remaining noisy segments are identified and removed through visual inspection. Bad channels are identified and corrected using spherical spline interpolation.

The clean EEG signals are then cut into 20-second segments where segments are non-overlapped. All segments from one patient stay in the same fold, and we compute the metrics per patient rather than per segment. To prevent data leakage, all preprocessing steps, including z-score normalization, are applied independently within each cross-validation fold using only the training data of that fold. These segments are the input to the TF image generation pipeline.

Each TF method converts one segment of one channel into one image. The primary cohort gives 4,142 TF images from 15 patients. The secondary cohort gives 12,122 TF images from 46 patients. This is about 276 images per patient in the primary cohort and 263 in the secondary cohort. Each patient contributes 19 EEG channels across about 14 retained segments. The primary cohort has 2,565 responder images and 1,577 non-responder images.
}
\begin{figure} [t]
\centering
\vskip\baselineskip
    \begin{subfigure}{0.45\textwidth}
        \centering
        \includegraphics[width=\linewidth]{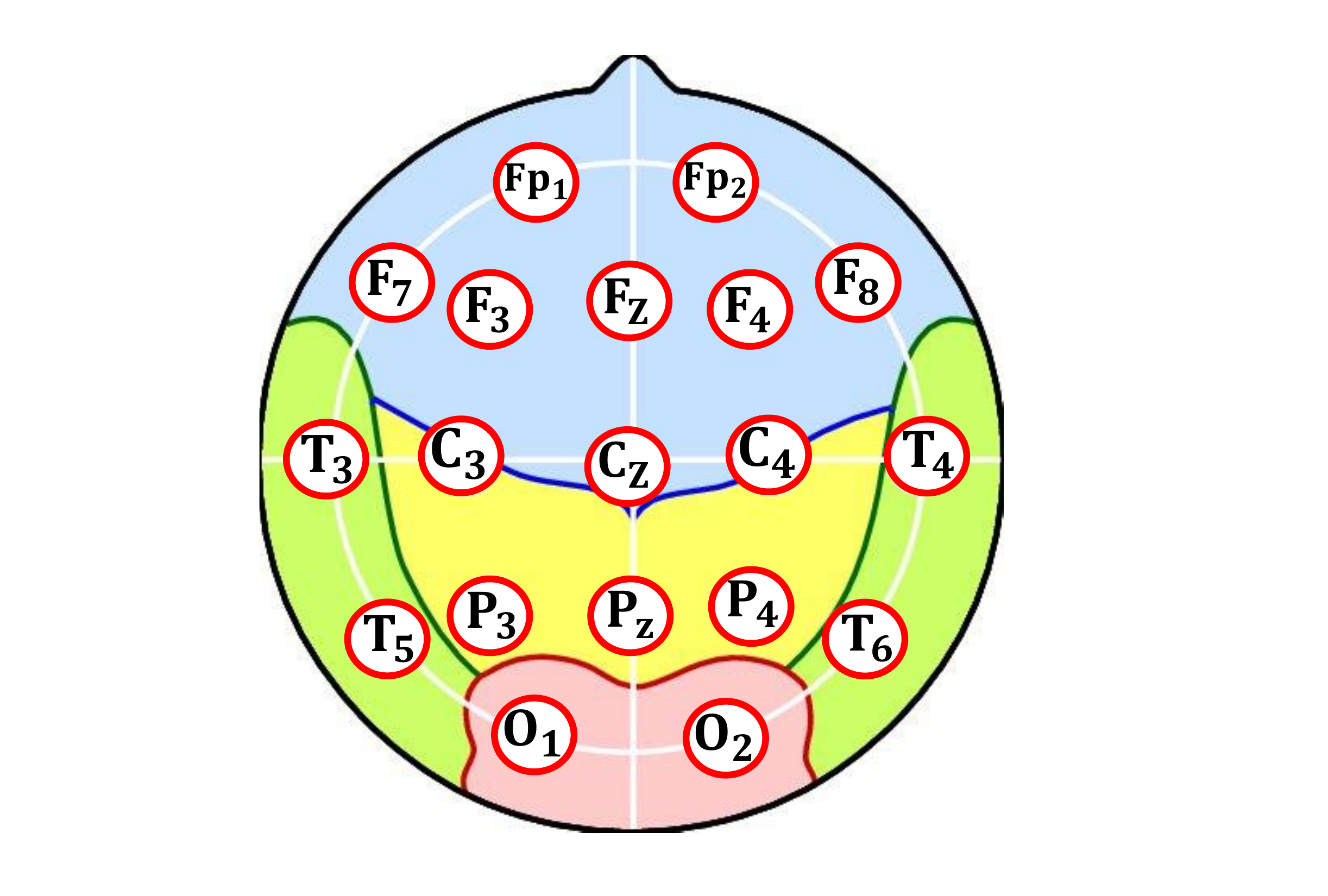}
    \end{subfigure}%
    \begin{subfigure}{0.45\textwidth}
        \centering
        \includegraphics[width=\linewidth]{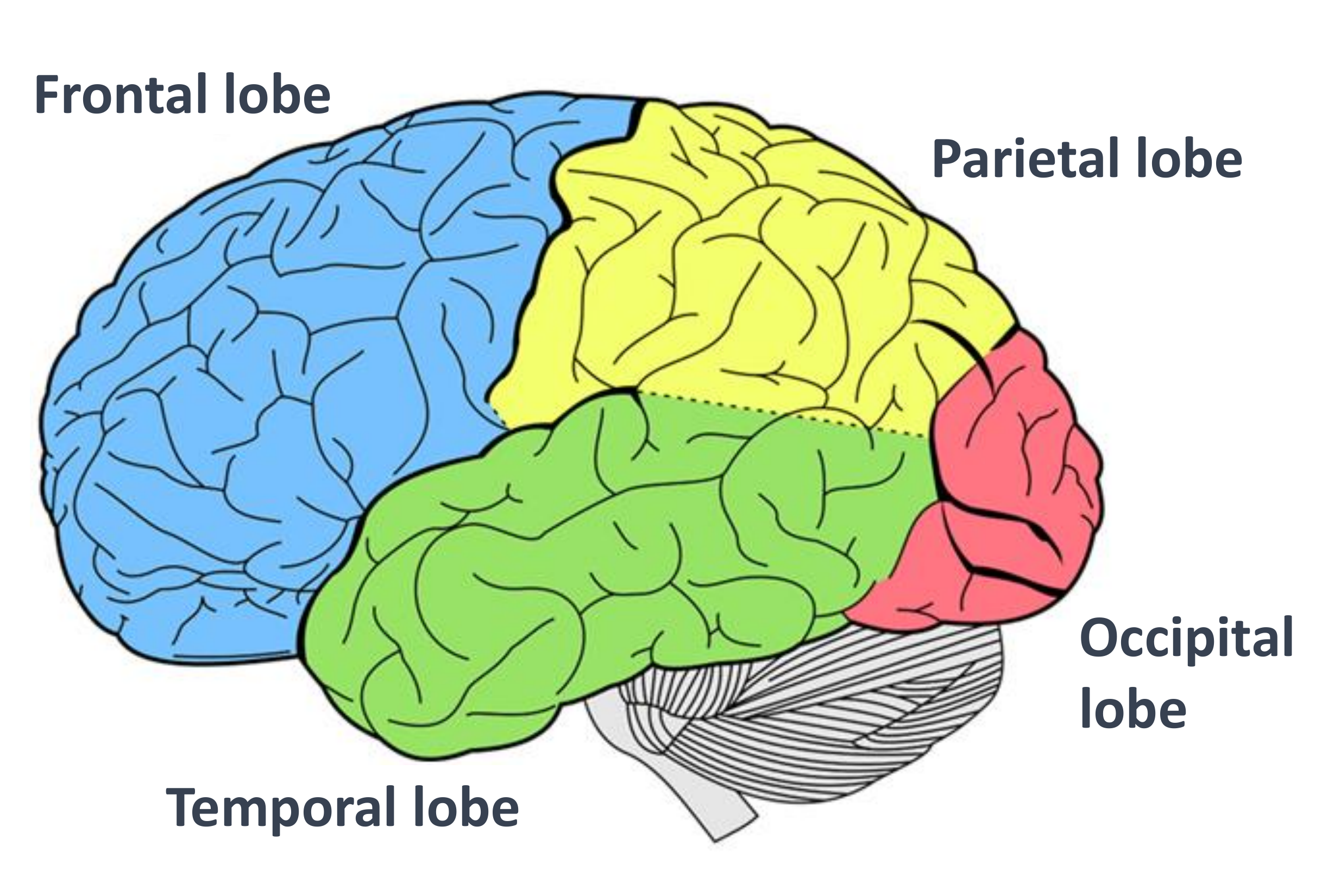}
    \end{subfigure}
    \caption{The location of channels in 10-20 system (left) and the brain lobes (right).}
    \label{Fig:channel-brain}
\end{figure}
\subsection{Time-frequency method fusion}
We use four different time-frequency methods for each patient to identify responders and non-responders. Each time-frequency method provides different types of features for extraction. In this paper, we introduce the fusion of two different time-frequency methods for more robust feature extraction: montage and blending. Montage places two different images from two different time-frequency methods side by side. Blending overlays two different images from two different time-frequency methods using alpha blending. This allows our CNN model to extract more complex features and learn more effectively. From our experiments in Table~\ref{tab:Performance_of_CNN_Independent} and Table~\ref{tab:pretrained_vs_cnn}, the CWT time-frequency method yields the highest accuracy among all methods. Our model extracts the most features from the CWT time-frequency representation. Therefore, we fuse CWT with other time-frequency methods using two different fusion techniques. We generate CWT\_DWT, CWT\_ST, and CWT\_SSWE using both montage and blending.
\subsubsection{Fused CWT and ST}
\begin{figure}
    \centering
    \begin{subfigure}{0.60\textwidth}
        \centering
        \includegraphics[width=\linewidth]{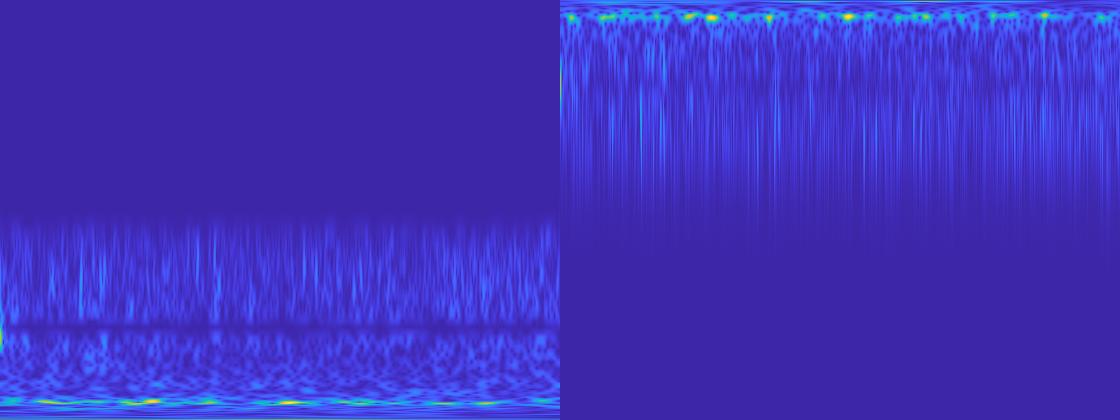}
        \caption{Montage}
        \label{rTMS_CWT_ST_R_montage}
    \end{subfigure}%
    \hspace{3 em} 
    \begin{subfigure}{0.30\textwidth}
        \centering
        \includegraphics[width=\linewidth]{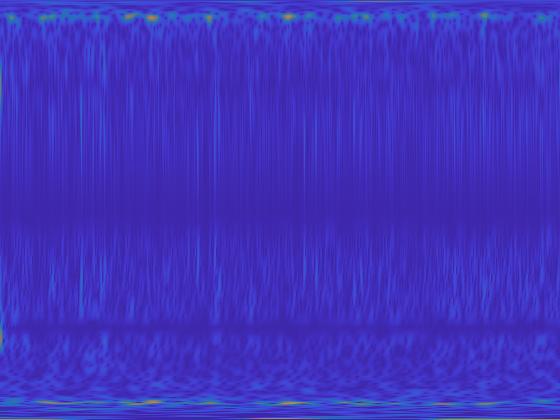}
        \caption{Blend}
        \label{rTMS_CWT_ST_R_blend}
    \end{subfigure}
    \caption{Sample Fused CWT\_ST Time Frequency Images of Responder}
    \label{rTMS_CWT_ST_fused_images}
\end{figure}

\textbf{Continuous Wavelet Transform (CWT)} is a widely used technique in signal processing that allows the simultaneous analysis of signals in both time and frequency domains. It is particularly effective for examining non-stationary signals within the time-frequency domain. The CWT works by convolving the signal with a set of wavelets that differ in scale and position, producing a time-frequency representation that accurately highlights localized features. It captures both low- and high-frequency components of a signal, making it well suited for analyzing signals with varying frequency content.
The \textbf{S-Transform (ST)} technique is a method for time-frequency analysis that examines how a signal changes over time at a single observation location. Unlike conventional methods, ST directly evaluates temporal variations in signal characteristics within the time domain. This makes it especially useful for analyzing localized or short-duration events. Since it maintains temporal resolution, it enables accurate identification of signal fluctuations without relying on complex transformations.
To fuse the CWT and ST time-frequency methods, we first transform EEG signals into CWT and ST-based time-frequency images. A precise setting is applied for both CWT and ST during generating fragments. For example, the signal data is divided into 20-second segments and the sampling rate is set at 500 Hz, ensuring high temporal resolution in frequency analysis. CWT and ST are applied to each segment, producing time-frequency images that provide a comprehensive representation of the signal's frequency distribution over time. After getting both the CWT and ST fragments, we combine them with two different fusion techniques, montage and blend. We then generate two different databases of images with CWT\_ST images shown in Fig.~\ref{rTMS_CWT_ST_fused_images}.
\subsubsection{Fused CWT and DWT}
\begin{figure}
    \centering
    \begin{subfigure}{0.60\textwidth}
        \centering
        \includegraphics[width=\linewidth]{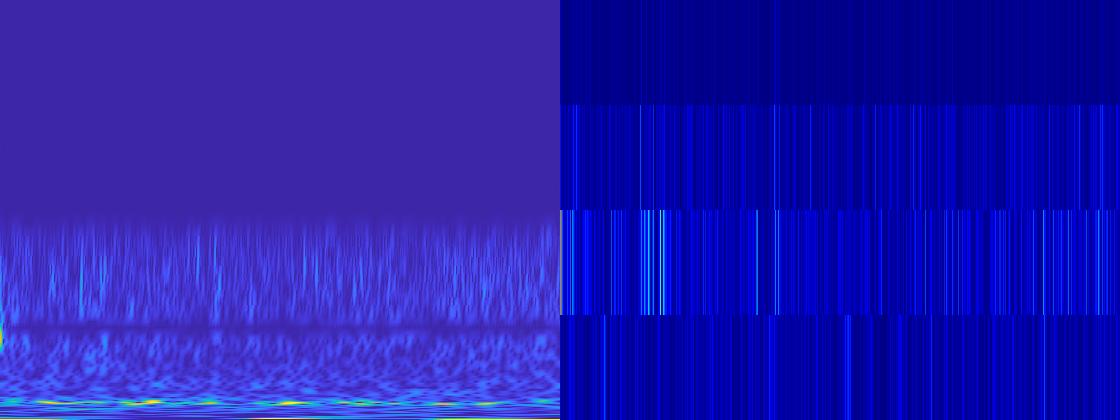}
        \caption{Montage}
        \label{rTMS_CWT_DWT_R_montage}
    \end{subfigure}%
    \hspace{3 em} 
    \begin{subfigure}{0.30\textwidth}
        \centering
        \includegraphics[width=\linewidth]{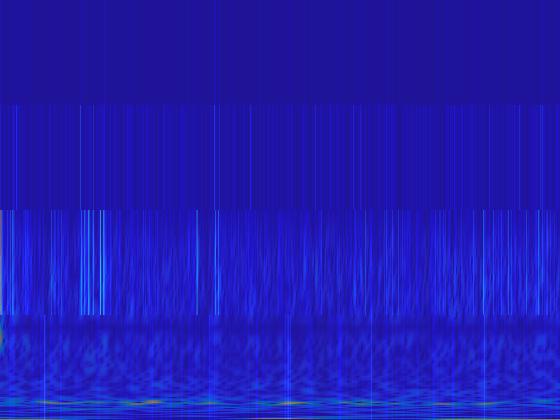}
        \caption{Blend}
        \label{rTMS_CWT_DWT_R_blend}
    \end{subfigure}
    \caption{Sample Fused CWT\_DWT Time Frequency Images of Responder}
    \label{rTMS_CWT_DWT_fused_images}
\end{figure}

The \textbf{Discrete Wavelet Transform (DWT)} is a technique for analyzing signals in both time and frequency domains. It splits a signal into multiple frequency bands, each with a distinct resolution. Unlike the CWT, which functions across continuous scales, the DWT works on discrete scales. This is done by repeatedly breaking down the signal using both high-pass and low-pass filters, creating a multi-level representation. Such decomposition helps in detecting sudden variations and transient components in signals. Although DWT reduces time localization at elevated frequencies to enhance computational efficiency, it remains an effective method for identifying key signal features. It is widely applied in tasks like signal compression, feature extraction, and noise reduction.
After generating CWT frames, to transform EEG signals into time-frequency images using the DWT, the signal data is divided into 20-second segments. A sampling rate of 500 Hz is selected to ensure high-resolution data collection for precise decomposition. Each segment undergoes DWT using a multi-level decomposition technique, allowing the extraction of time-frequency features at different wavelet levels. These extracted features are then represented as scalogram images. We then combine both CWT and DWT frames with two different fusion techniques, montage and blend, as shown in Fig.~\ref{rTMS_CWT_DWT_fused_images}.
\subsubsection{Fused CWT and SSWE}
\begin{figure}
    \centering
    \begin{subfigure}{0.60\textwidth}
        \centering
        \includegraphics[width=\linewidth]{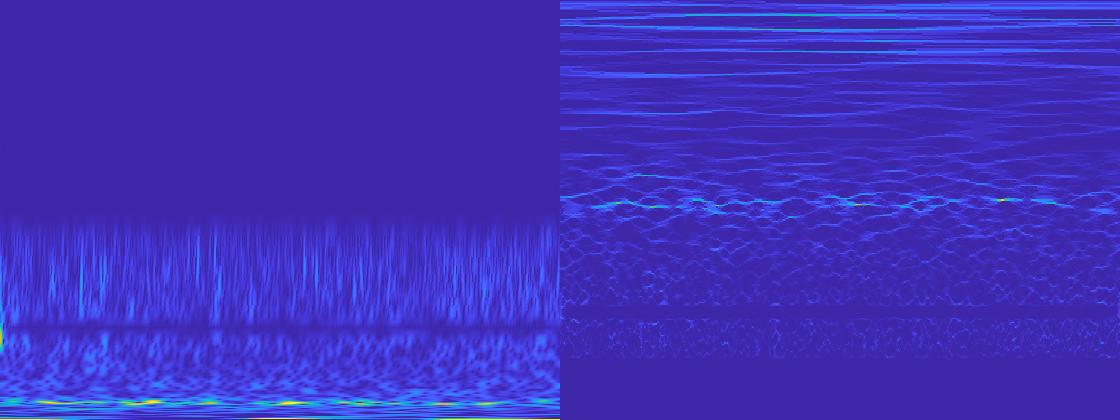}
        \caption{Montage}
        \label{rTMS_CWT_SSWE_R_montage}
    \end{subfigure}%
    \hspace{3 em} 
    \begin{subfigure}{0.30\textwidth}
        \centering
        \includegraphics[width=\linewidth]{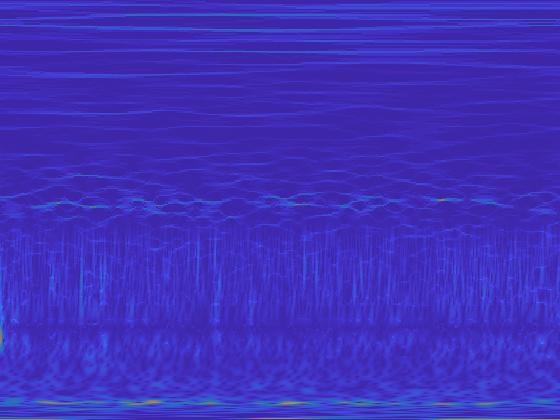}
        \caption{Blend}
        \label{rTMS_CWT_SSWE_R_blend}
    \end{subfigure}
    \caption{Sample Fused CWT\_SSWE Time Frequency Images of Responder}
    \label{rTMS_CWT_SSWE_fused_images}
\end{figure}

The \textbf{Wavelet Synchrosqueezing Transform (SSWE)} is a powerful method used to study signals that contain several changing patterns or frequencies. Compared to regular wavelet transforms, SSWE gives a clearer picture by reducing the blurring that usually happens in time-frequency analysis. It works by shifting energy in the frequency direction while keeping the timing details of the signal intact. To use SSWE, we first select a special type of wavelet, then calculate the CWT, and finally use a phase-based technique to find the signal's exact frequency at each moment. By moving energy based on the phase, SSWE makes the signal's time-frequency display sharper and easier to read.
After creating CWT frames, we turn EEG signals into SSWE-based time-frequency images. The recordings are first divided into smaller parts, each lasting 20 seconds. These segments are sampled at 500 Hz. Before applying any transformations, the signals are normalized to keep the amplitude levels consistent across all samples. The SSWE method is then used on each segment to produce a high-resolution time-frequency view. After generating both CWT and SSWE frames, we combine both time-frequency methods with two fused techniques, montage and blend, as shown in Fig.~\ref{rTMS_CWT_SSWE_fused_images}.
\subsection{Data Splitting and Cross-Validation Setup}
Using four different time-frequency methods, we convert EEG signals into different time-frequency images. This results in four different datasets of images. The dataset used in our study comprises images derived from individual and combined CWT, DWT, SSWE, and ST EEG signals. Each dataset is organized into two classes: Responder and Non-Responder. The raw datasets are split into 10 folds for cross-validation, with the splits stored in a separate directory.
A 10-fold cross-validation strategy is used in both the proposed CNN and the pretrained models to evaluate model performance comprehensively. We use KFold from the Scikit-learn library to generate training and testing splits for each fold. To ensure reproducibility, the random seed for shuffling during splitting is fixed (random\_state=42). Each fold takes a turn as the test set, while the remaining 9 folds form the training set. This process runs 10 times and uses 1 fold for testing each time.
\textcolor{black}{This 10-fold split is applied at the segment level and is used only for the experiments in Sections 4.1 to 4.4. Segments from the same patient can fall in different folds, so a patient can appear in both training and testing. All subject-level experiments in Section 4.5 use a separate stratified group cross-validation protocol, described in Section 4.5.1, where no patient appears in both training and testing.}
\subsection{Proposed CNN}
Our proposed model is a custom CNN designed for binary classification. The architecture includes an Input Layer which accepts images of size (224 $\times$ 224 $\times$ 3). We use three convolutional layers with 8, 16, and 32 filters, each using (3$\times$3) kernels and ReLU activation. These layers are designed to extract low- to high-level features from the input images. After each convolutional layer we use Batch Normalization to stabilize learning and improve convergence. AveragePooling2D is applied with a pool size of (2, 2) to downsample the feature maps. To convert the feature maps into a 1D vector we use a flatten layer. A dropout rate of 25\% is used to reduce overfitting by randomly deactivating neurons during training. As a fully connected layer, we use a dense layer which is a single neuron with a sigmoid activation function for binary classification. Our model is compiled using the Adam optimizer with a learning rate of 0.0001 and Binary cross-entropy is used as the loss function, and accuracy is the evaluation metric. To prevent overfitting, early stopping is used with a patience of five as a regularization technique. In Fig~\ref{blockD} the architecture of our proposed CNN is shown.
\subsection{TL Architecture}
In this study, several state-of-the-art transfer learning models are used to classify EEG signals represented as time-frequency images. The transfer learning approach uses pre-trained models originally trained on the large-scale ImageNet dataset and fine-tunes them for binary classification of EEG signals (Responder vs. Non-Responder).
Each model follows a consistent structure with a pre-trained base and custom layers added for binary classification. The pre-trained weights from the ImageNet dataset are loaded. The base model's layers are initially frozen to retain the learned features. We implement a Global Average Pooling layer to reduce the feature maps to a single value per channel, minimizing the number of trainable parameters. Batch Normalization is used to stabilize and accelerate training by normalizing activations. A fully connected dense layer with 256 neurons and ReLU activation is used for feature learning. A dropout rate of 30--50\% is applied to reduce overfitting. Then a single neuron output layer with a sigmoid activation function is used for binary classification. This architecture is consistent across all transfer learning models, with only the base model (VGG16, Xception, ResNet152V2, MobileNetV2, and DenseNet201) being replaced in each experiment. Throughout our model, Adam optimizer with a learning rate of 0.0001 and early stopping are used to prevent overfitting.
\subsection{Performance Evaluation}
To comprehensively evaluate the predictive efficacy of the proposed classification framework, several standard evaluation metrics are computed from the model's confusion matrix. Given that clinical datasets are frequently characterized by class imbalances, relying solely on overall classification accuracy can yield overly optimistic or misleading conclusions. Therefore, a multi-metric paradigm comprising accuracy, precision, recall (sensitivity), specificity, and F1-score is used to rigorously characterize model behavior. These metrics are mathematically defined as follows:
\begin{equation}
\text{Accuracy} = \frac{\text{TP} + \text{TN}}{\text{TP} + \text{FP} + \text{TN} + \text{FN}}
\end{equation}
\begin{equation}
\text{Precision} = \frac{\text{TP}}{\text{TP} + \text{FP}}
\end{equation}
\begin{equation}
\text{Recall (Sensitivity)} = \frac{\text{TP}}{\text{TP} + \text{FN}}
\end{equation}
\begin{equation}
\text{Specificity} = \frac{\text{TN}}{\text{TN} + \text{FP}}
\end{equation}
\noindent where $\text{TP}$, $\text{TN}$, $\text{FP}$, and $\text{FN}$ denote the number of true positives (correctly identified responders), true negatives (correctly identified non-responders), false positives (non-responders misclassified as responders), and false negatives (responders misclassified as non-responders), respectively.
To assess performance balance, the $F_{1}\text{-score}$ is computed as the harmonic mean of precision and recall:
\begin{equation}
F_{1}\text{-score} = 2 \cdot \frac{\text{Precision} \cdot \text{Recall}}{\text{Precision} + \text{Recall}}
\end{equation}

Beyond discrete classification thresholds, the diagnostic discriminative capacity of the model across varying decision boundaries is quantified using the Receiver Operating Characteristic (ROC) curve and its corresponding Area Under the Curve (AUC).

\textcolor{black}{The metrics above evaluate the decisions the model makes after a threshold is applied. They do not evaluate the probabilities it produces before that threshold. To assess calibration we report the Brier score, which is the mean squared difference between the predicted probability and the observed outcome:}
\begin{equation}
\text{Brier} = \frac{1}{N}\sum_{i=1}^{N}\left(p_i - y_i\right)^2
\end{equation}
\noindent \textcolor{black}{where $p_i$ is the predicted probability for patient $i$ and $y_i$ is that patient's outcome, coded 1 for responder and 0 for non-responder. Lower values are better. The Brier score has a natural reference point. A model that ignores the input and predicts the base rate for every patient scores 0.24 on the primary cohort, where the base rate is 9 of 15, and 0.25 on the secondary cohort, where it is 23 of 46. We report the Brier score for every configuration in Tables~\ref{tab:sub_sm_gen}, \ref{tab:dataset_modalities_grouped}, \ref{tab:dataset2_modalities} and \ref{tab:dataset2_modalities_grouped}.}

\textcolor{black}{To address the constraints of compact clinical sample sizes, all subject-level experiments are repeated across several randomized initialization seeds. We report performance as $\text{mean} \pm \text{standard deviation}$ across seeds. We also report a 95\% confidence interval for the AUC. The interval comes from a bootstrap that resamples patients with replacement and recomputes the AUC 2,000 times. We take the 2.5th and 97.5th percentiles of the resulting distribution as the interval. We resample patients and not segments, because the patient is the unit of analysis and segments from one patient are not independent of each other.}

{\color{black}
Confidence intervals alone do not establish whether the measured performance is real. We therefore add a patient-level permutation test. We shuffle the responder and non-responder labels across whole patients, so that every segment of one patient carries the same shuffled label, and we then rerun the complete cross-validation pipeline. Repeating this many times builds the distribution of performance that the pipeline produces when the labels carry no information. We compare the observed value against that distribution and report the $p$ value as the fraction of null values that reach or exceed the observed value, with the standard add-one correction:
\begin{equation}
p = \frac{\#\{\text{null} \geq \text{observed}\} + 1}{n_{\text{perm}} + 1}
\end{equation}
\noindent We shuffle at the patient level and not at the segment level. Shuffling single segments would leave the patient structure intact, which gives an easier null and inflates significance. We treat $p < 0.05$ as the threshold. Where several configurations are evaluated on the same cohort we also report the Bonferroni-corrected threshold and state how many configurations were run. We report the outcome of every test we run, including the tests that do not reach significance.
}
\section{Results}
\subsection{Preliminary experiment}
Since we have four different individual time-frequency methods, we need to identify the best time-frequency method to fuse with other time-frequency methods. In our first experiment, we use the proposed custom CNN to find out the best time-frequency method for predicting the outcome of rTMS depression therapy. The performance of our proposed custom CNN model is evaluated across four time-frequency methods: CWT, DWT, ST, and SSWE. Table~\ref{tab:Performance_of_CNN_Independent} summarizes the classification metrics for both Non-Responder (NR) and Responder (R) classes.
Our experiments show that with our proposed CNN, the CWT time-frequency method achieves the highest overall performance, with an Accuracy of 98.67\% and an AUC of 98.52\%. This demonstrates the suitability of our proposed CNN for time-frequency representations using CWT.
\begin{table}
\centering
\resizebox{0.8\textwidth}{!}{%
\begin{tabular}{|c|cc|cc|cc|cc|c|c|}
\hline
\multirow{2}{*}{Dataset} & \multicolumn{2}{c|}{Precision \%} & \multicolumn{2}{c|}{Recall \%} & \multicolumn{2}{c|}{Specificity \%} & \multicolumn{2}{c|}{$F_{\text{score}}$ \%} & \multirow{2}{*}{AUC} & \multirow{2}{*} {Accuracy \%}   \\
\cline{2-9}
  & NR & R & NR & R & NR & R & NR & R &  & \\
\hline
DWT & 90 & 90 & 84 & 94 & 90 & 88 & 87 & 92 & 88.99 & 90.22 \\
ST & 98 & 98 & 97 & 99 & 98 & 97 & 97 & 98 & 97.76 & 97.78 \\
SSWE & 98 & 98 & 96 & 99 & 98 & 97 & 97 & 98 & 97.76 & 97.97 \\
\textbf{CWT} & \textbf{99} & \textbf{99} & \textbf{98} & \textbf{99} & \textbf{99} & \textbf{98} & \textbf{98} & \textbf{99} & \textbf{98.52} & \textbf{98.67} \\
\hline
\end{tabular}
 }
 \caption{Performance measures of our Proposed CNN using four time-frequency methods on Primary Dataset}\label{tab:Performance_of_CNN_Independent}
\end{table}
As shown in Table~\ref{tab:Performance_of_CNN_Independent}, the performance of our CNN with CWT time-frequency images is superior. We conclude that CWT images perform significantly better for effective feature extraction compared to other time-frequency methods, DWT, SSWE, and ST. For further experiments, we fuse CWT with other time-frequency methods to make new fused datasets.
\subsection{Pretrained architectures}
In our second experiment, we use CWT time-frequency images to compare with several state-of-the-art pretrained models, including VGG16, Xception, ResNet152V2, MobileNetV2, and DenseNet201. We conduct this experiment to see how state-of-the-art models perform on CWT time-frequency images and to use as further comparison. The results are summarized in Table~\ref{tab:pretrained_vs_cnn}.
\begin{table}
\centering
\resizebox{0.8\textwidth}{!}{%
\begin{tabular}{|c|cc|cc|cc|cc|c|c|}
\hline
\multirow{2}{*}{Model} & \multicolumn{2}{c|}{Precision \%} & \multicolumn{2}{c|}{Recall \%} & \multicolumn{2}{c|}{F1-score \%} & \multicolumn{2}{c|}{Specificity \%} & \multirow{2}{*}{AUC} & \multirow{2}{*}{Accuracy \%} \\
\cline{2-9}
 & NR & R & NR & R & NR & R & NR & R & & \\
\hline
VGG16 & 85 & 83 & 70 & 93 & 77 & 88 & 85 & 83 & 81.24 & 83.94 \\
Xception & 87 & 87 & 77 & 93 & 82 & 90 & 87 & 87 & 85.10 & 87.01 \\
ResNet152V2 & 90 & 87 & 76 & 95 & 83 & 91 & 90 & 87 & 85.60 & 87.86 \\
MobileNetV2 & 91 & 90 & 83 & 95 & 86 & 92 & 91 & 90 & 88.69 & 90.13 \\
DenseNet201 & 91 & 91 & 84 & 95 & 87 & 93 & 93 & 90 & 89.44 & 90.78 \\
\textbf{Proposed CNN} & \textbf{99} & \textbf{99} & \textbf{98} & \textbf{99} & \textbf{98} & \textbf{99} & \textbf{99} & \textbf{98} & \textbf{98.52} & \textbf{98.67} \\
\hline
\end{tabular}%
}
\caption{Performance of CWT time-frequency method with Proposed CNN VS Pre-trained models}\label{tab:pretrained_vs_cnn}
\end{table}
\subsection{Fused time frequency methods}
\begin{figure}[ht]
\centering
\includegraphics[width=0.99\textwidth]{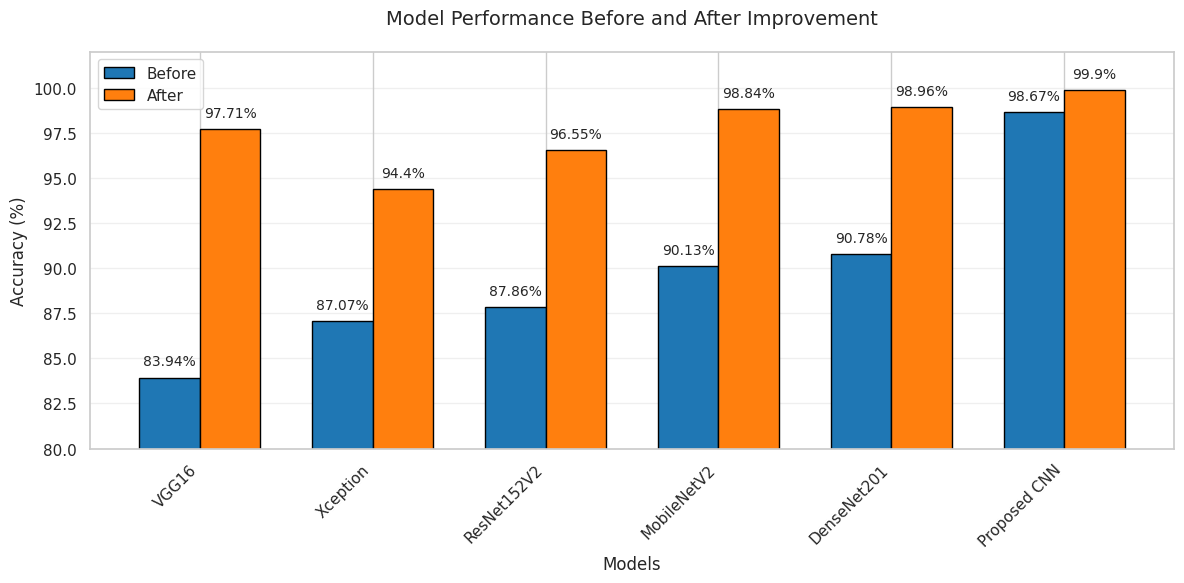}
\caption{\textcolor{black}{Performance comparison of individual CWT (Before) and fused Montage CWT\_ST (After), showing accuracy improvement across all models.}}
\label{beforeAfter}
\end{figure}

\begin{figure}[ht]
\fontsize{9pt}{11pt}\selectfont
    \centering
    \begin{subfigure}{0.32\textwidth}
        \centering
        \includegraphics[width=\linewidth]{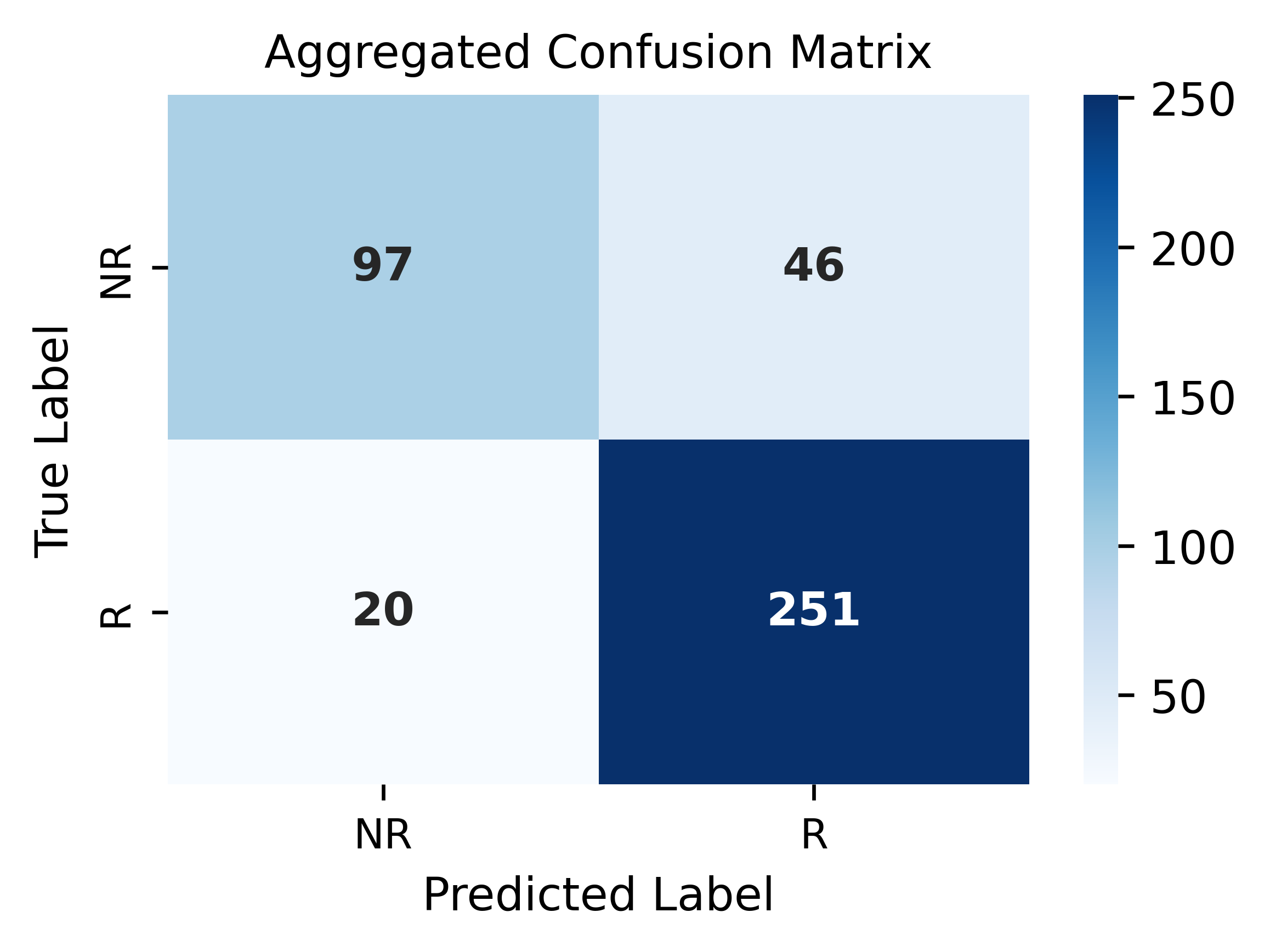}
        \caption{VGG16's Confusion Matrix \\ (CWT)}
        \label{fig:vgg16_cwt_cm}
    \end{subfigure}%
    \hfill
    \begin{subfigure}{0.32\textwidth}
        \centering
        \includegraphics[width=\linewidth]{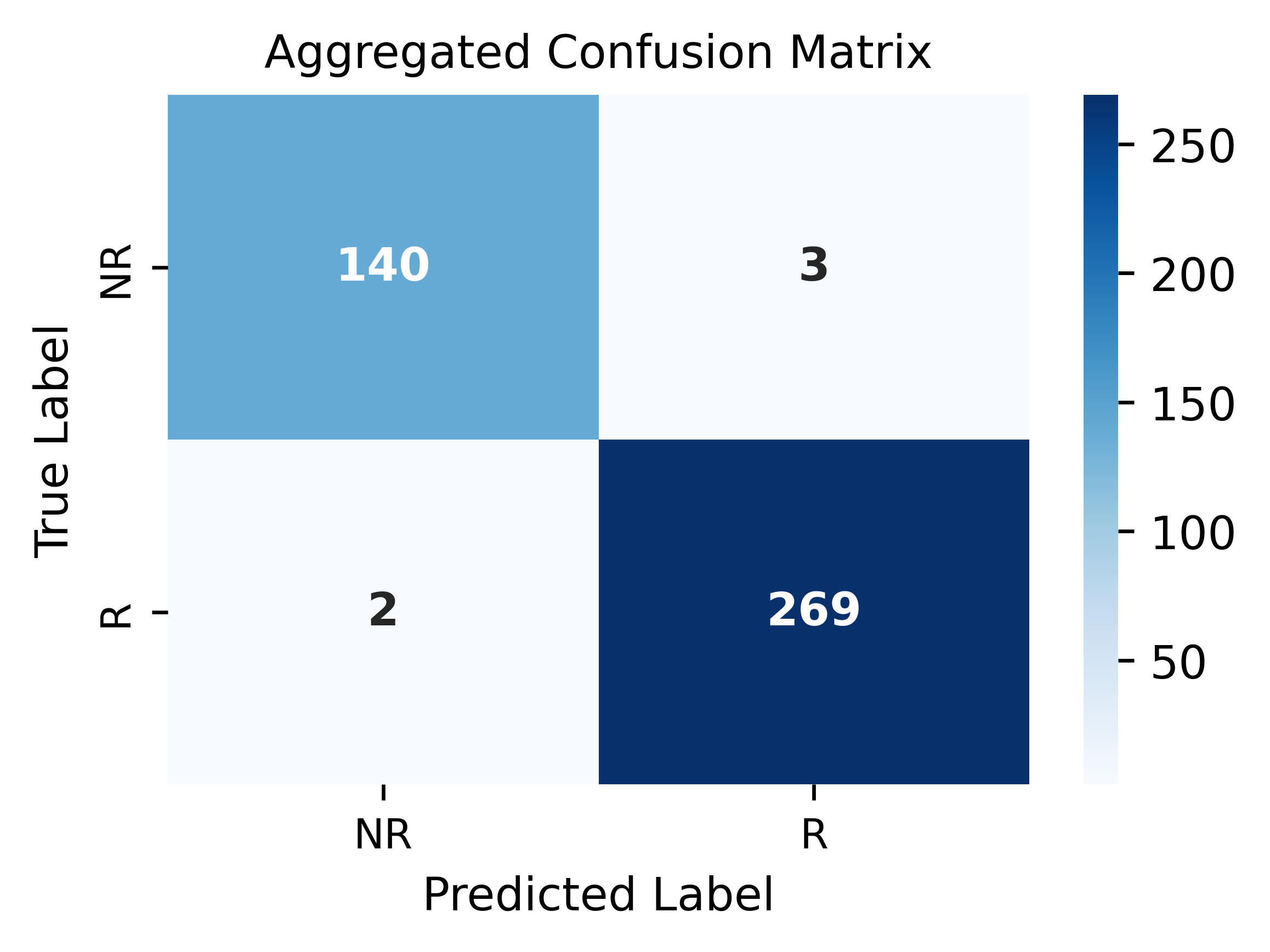}
        \caption{VGG16's Confusion Matrix (CWT\_ST)}
        \label{fig:vgg16_cwtst_cm}
    \end{subfigure}
    \hfill
    \begin{subfigure}{0.32\textwidth}
        \centering
        \includegraphics[width=\linewidth]{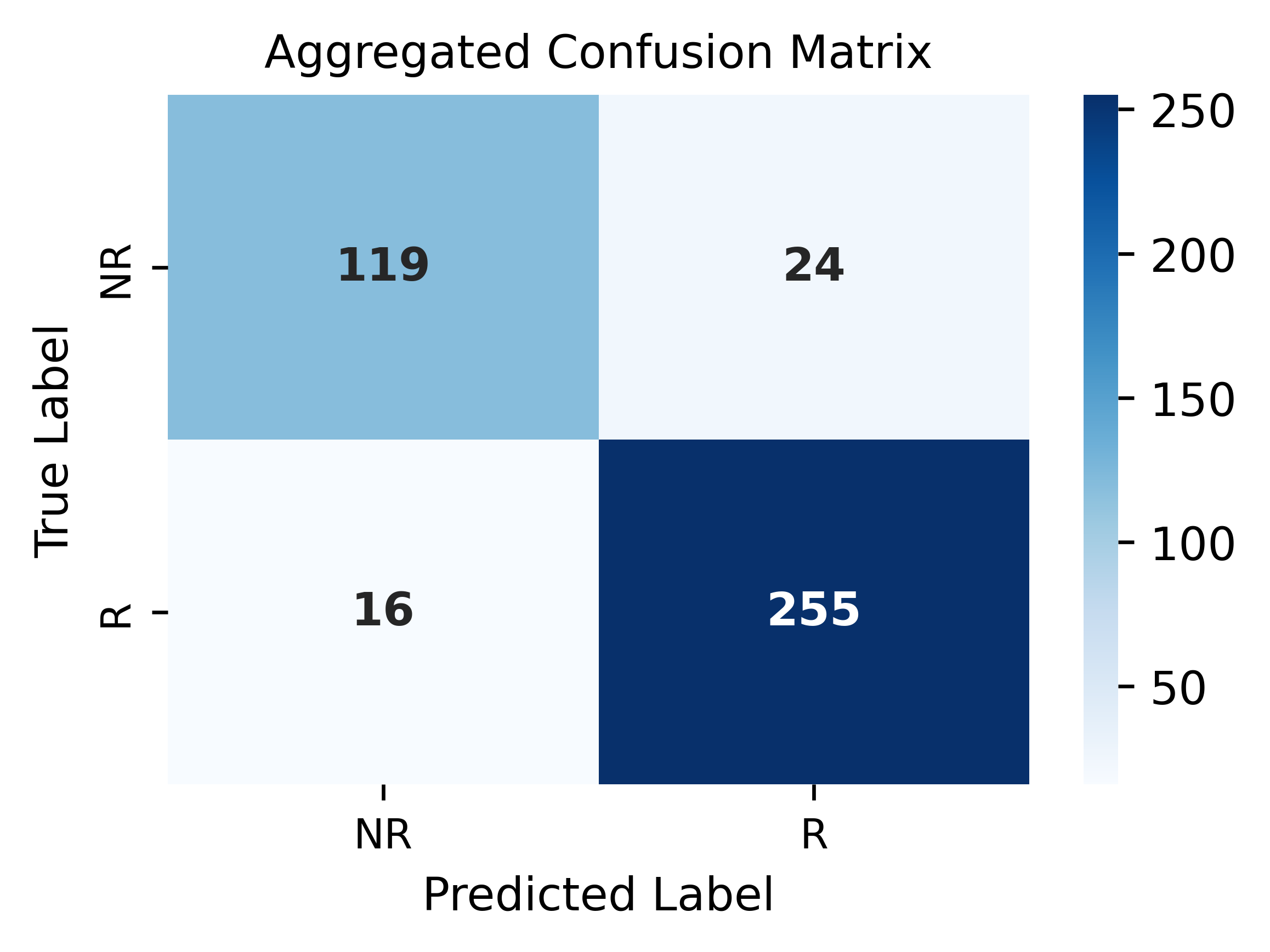}
        \caption{DenseNet201's Confusion Matrix (CWT)}
        \label{fig:densenet_cwt_cm}
    \end{subfigure}%
    \hspace{3em}
    \begin{subfigure}{0.32\textwidth}
        \centering
        \includegraphics[width=\linewidth]{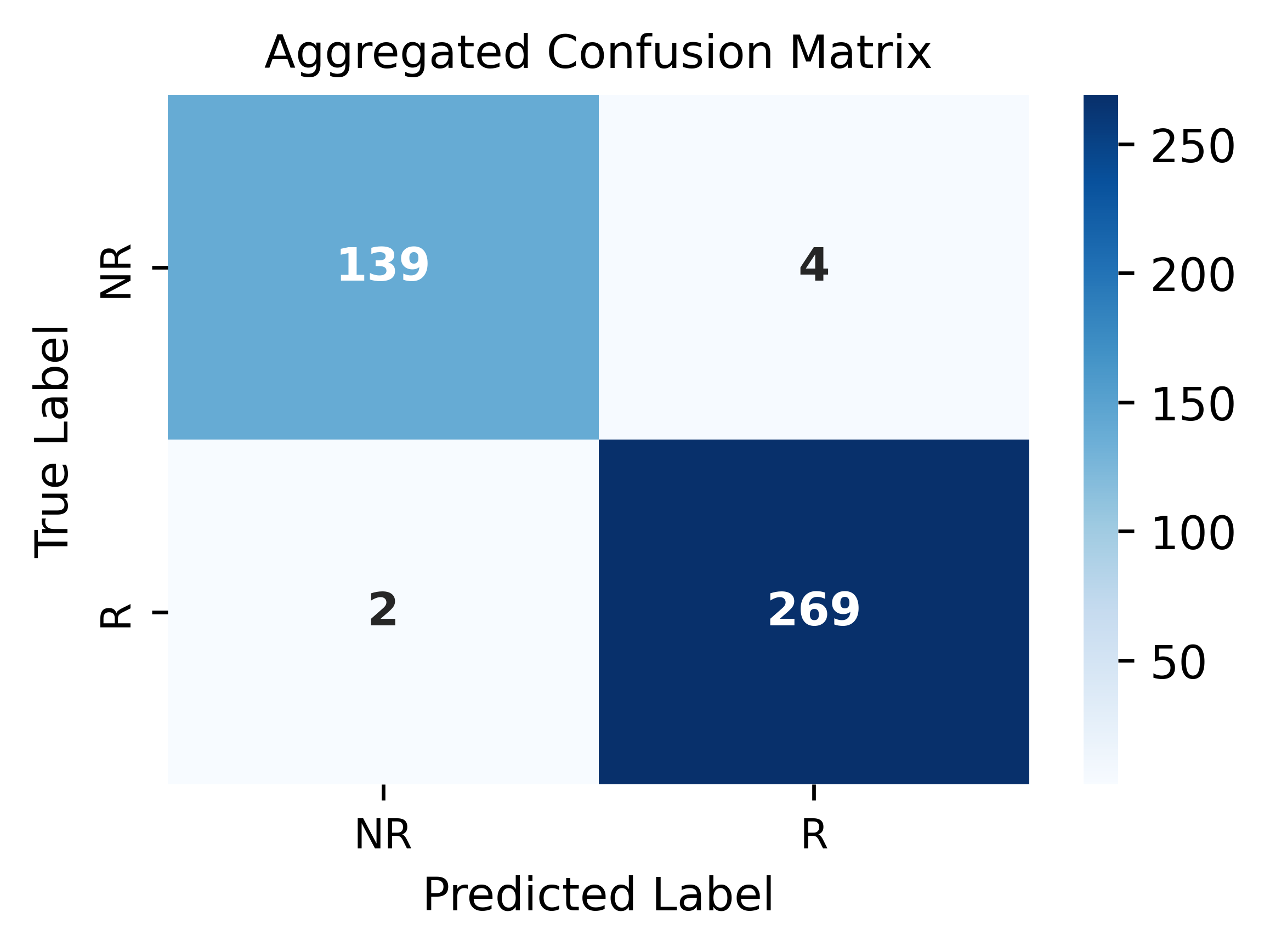}
        \caption{DenseNet201's Confusion Matrix (CWT\_ST)}
        \label{fig:densenet_cwtst_cm}
    \end{subfigure}
     \hfill
    \begin{subfigure}{0.32\textwidth}
        \centering
        \includegraphics[width=\linewidth]{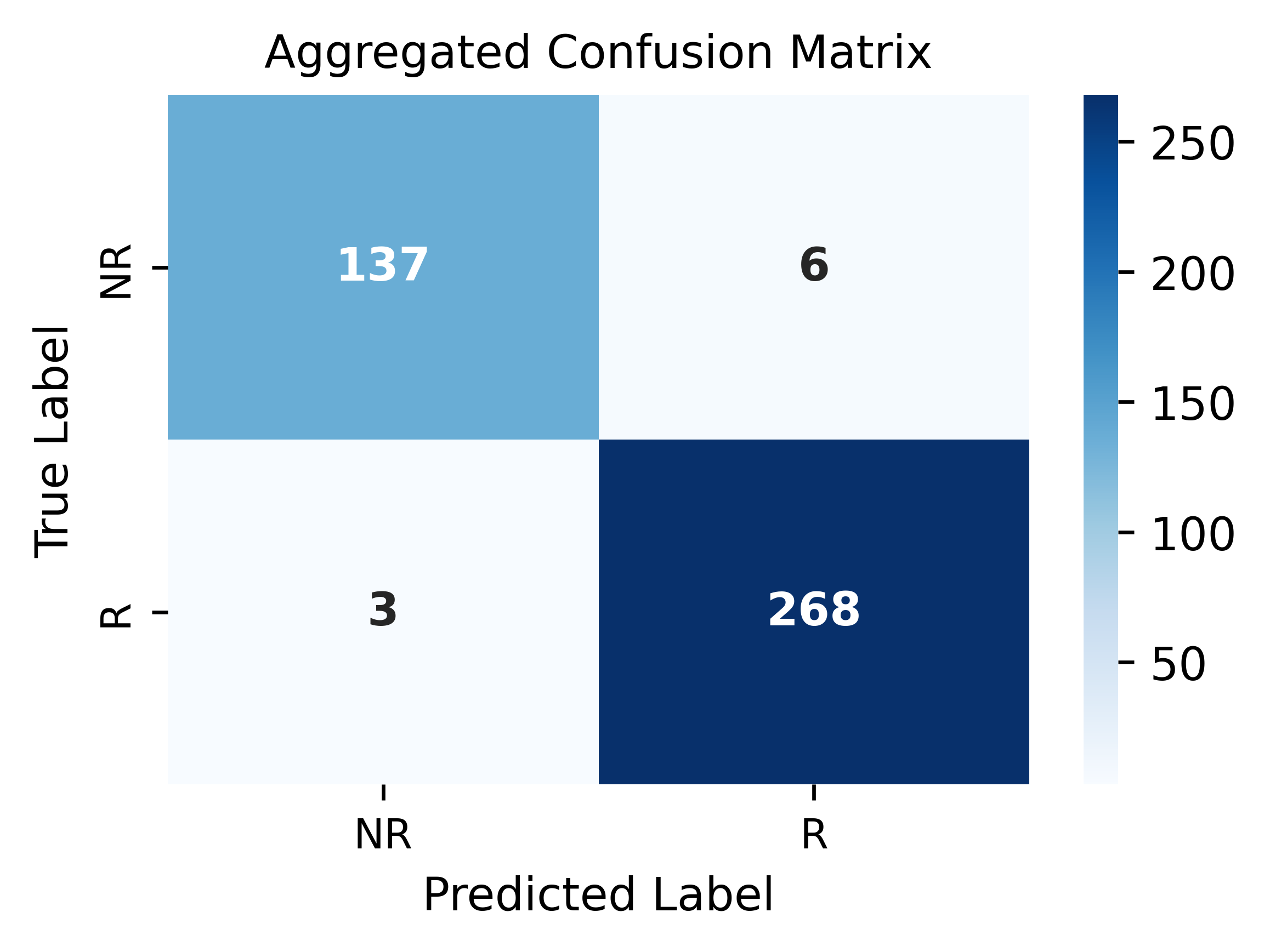}
        \caption{Custom CNN's Confusion Matrix (CWT)}
        \label{fig:cnn_cwt_cm}
    \end{subfigure}%
    \hfill
    \begin{subfigure}{0.32\textwidth}
        \centering
        \includegraphics[width=\linewidth]{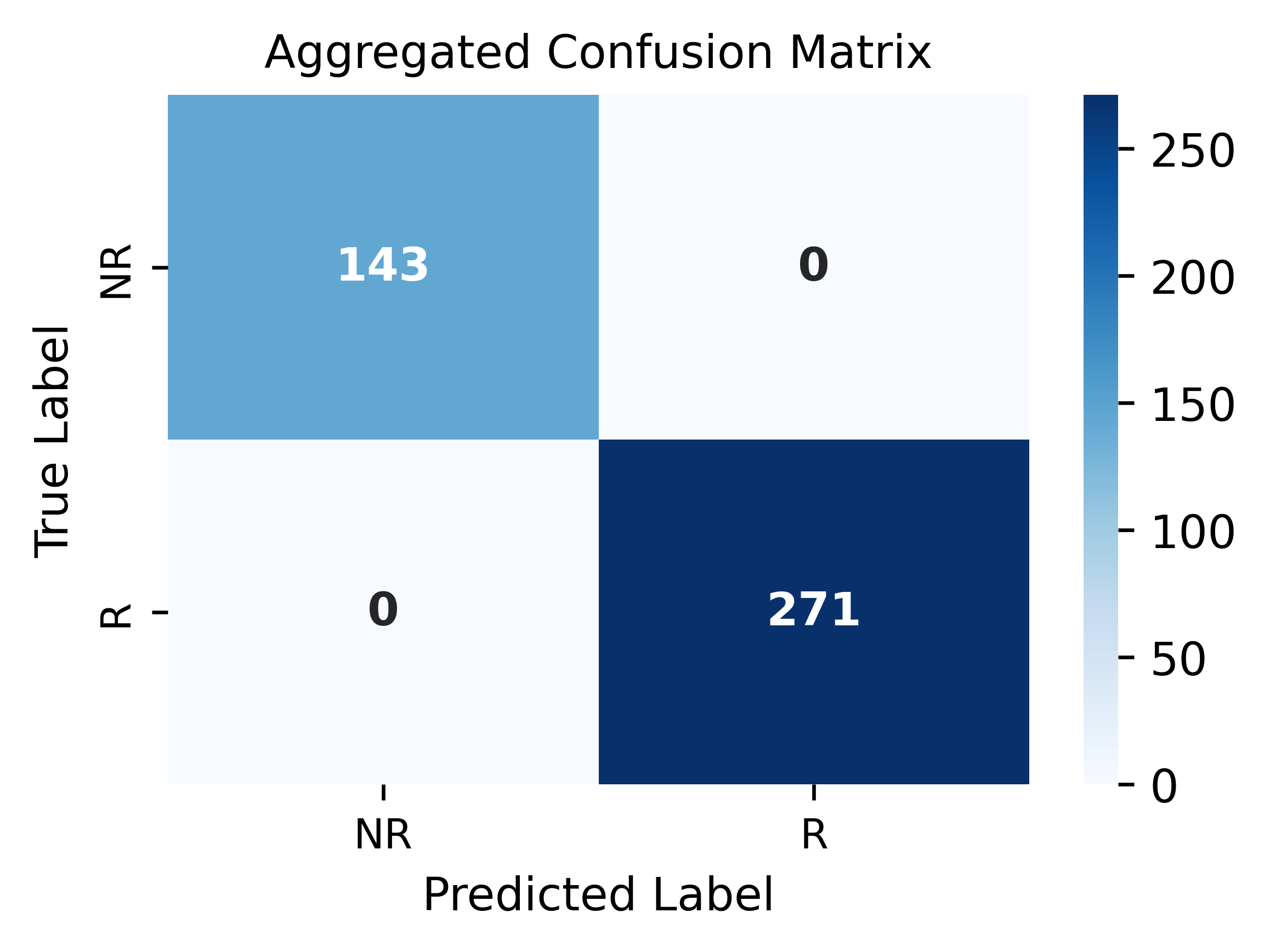}
        \caption{Custom CNN's Confusion Matrix (CWT\_ST)}
        \label{fig:cnn_cwtst_cm}
    \end{subfigure}
    \caption{Confusion Matrix of Individual and Fused Time-Frequency Method Using VGG16, DenseNet201 \& Proposed CNN}
    \label{fig:combined_CM_figures}
\end{figure}

CWT time-frequency method achieves the highest accuracy in both the proposed CNN and other state-of-the-art pretrained models. This means that CWT holds the most features that we can extract and fuse with other time-frequency methods for more robust feature extraction. In our third experiment, we use two different fusion techniques: blending (overlays A and B using alpha blending) and montage (puts A and B next to each other in the same image). For each fusion technique, we fuse two different time-frequency methods and generate 3 different sources of fused images: CWT\_ST, CWT\_SSWE, and CWT\_DWT. In total, for two different fusion techniques, we have six different fused datasets of images. We use the same proposed CNN architecture to experiment with each of these fused time-frequency methods. Below, we provide a performance comparison of our proposed CNN on different fused methods in Table~\ref{tab:Performance_of_Fused_Methods}.
\begin{table}
\centering
 \resizebox{0.8\textwidth}{!}{%
\begin{tabular}{|c|cc|cc|cc|cc|c|c|}
\hline
\multirow{2}{*}{Fused} & \multicolumn{2}{c|}{Precision \%} & \multicolumn{2}{c|}{Recall \%} & \multicolumn{2}{c|}{Specificity \%} & \multicolumn{2}{c|}{$F_{\text{score}}$ \%} & \multirow{2}{*}{AUC} & \multirow{2}{*} {Accuracy \%} \\
\cline{2-9}
  & NR & R & NR & R & NR & R & NR & R &  & \\
\hline
\multicolumn{11}{|c|}{\textbf{Montage}} \\
\hline
CWT\_DWT & 96 & 97 & 94 & 98 & 96 & 97 & 95 & 97 & 96.05 & 96.48 \\
CWT\_SSWE & 98 & 99 & 98 & 99 & 98 & 99 & 98 & 99 & 98.22 & 98.31 \\
\textbf{CWT\_ST} & \textbf{1.00} & \textbf{1.00} & \textbf{1.00} & \textbf{1.00} & \textbf{1.00} & \textbf{1.00} & \textbf{1.00} & \textbf{1.00} & \textbf{99.91} & \textbf{99.90} \\
\hline
\multicolumn{11}{|c|}{\textbf{Blend}} \\
\hline
CWT\_DWT & 96 & 96 & 93 & 98 & 96 & 96 & 95 & 97 & 95.39 & 95.94 \\
CWT\_SSWE & 98 & 98 & 97 & 99 & 98 & 98 & 98 & 99 & 97.98 & 98.17 \\
\textbf{CWT\_ST} & \textbf{98} & \textbf{99} & \textbf{98} & \textbf{99} & \textbf{98} & \textbf{99} & \textbf{98} & \textbf{99} & \textbf{98.24} & \textbf{98.33} \\
\hline
\end{tabular}
}
\caption{Performance measures of Proposed CNN on fused methods}\label{tab:Performance_of_Fused_Methods}
\end{table}
With our proposed custom CNN, the CWT\_ST time-frequency method achieves the highest accuracy in both montage (99.90\%) and blending (98.33\%) fusion techniques. These results highlight the superior performance of the CWT\_ST method, particularly in the montage configuration, making it a robust method for MDD treatment response prediction. To avoid biases and to validate our custom CNN's result, we conduct our fourth set of experiments with state-of-the-art pretrained models, including VGG16, Xception, ResNet152V2, MobileNetV2, and DenseNet201 using our fused dataset. In Table~\ref{tab:Performance_of_Proposed_CNN_vs_Pretrained_Montage}, we provide a performance comparison between our proposed CNN and pretrained models on the montage CWT\_ST dataset.
\begin{table}
\centering
\resizebox{0.8\textwidth}{!}{%
\begin{tabular}{|c|cc|cc|cc|cc|c|c|}
\hline
\multirow{2}{*}{Model} & \multicolumn{2}{c|}{Precision \%} & \multicolumn{2}{c|}{Recall \%} & \multicolumn{2}{c|}{Specificity \%} & \multicolumn{2}{c|}{$F_{\text{score}}$ \%} & \multirow{2}{*}{AUC} & \multirow{2}{*}{Accuracy \%} \\
\cline{2-9}
  & NR & R & NR & R & NR & R & NR & R &  & \\
\hline
VGG16 & 98 & 97 & 96 & 99 & 99 & 96 & 97 & 98 & 97.32 & 97.71 \\
Xception & 95 & 94 & 90 & 97 & 97 & 90 & 92 & 96 & 93.55 & 94.40 \\
ResNet152V2 & 97 & 96 & 94 & 98 & 98 & 94 & 95 & 97 & 96.06 & 96.55 \\
MobileNetV2 & 99 & 99 & 98 & 99 & 99 & 98 & 98 & 99 & 98.64 & 98.84 \\
DenseNet201 & 99 & 99 & 98 & 99 & 99 & 98 & 99 & 99 & 98.81 & 98.96 \\
\textbf{Proposed CNN} & \textbf{1.00} & \textbf{1.00} & \textbf{1.00} & \textbf{1.00} & \textbf{1.00} & \textbf{1.00} & \textbf{1.00} & \textbf{1.00} & \textbf{99.91} & \textbf{99.90} \\
\hline
\end{tabular}
}
\caption{Performance comparison of Proposed CNN vs Pretrained models on CWT\_ST montage fusing technique}\label{tab:Performance_of_Proposed_CNN_vs_Pretrained_Montage}
\end{table}
Table~\ref{tab:Performance_of_Proposed_CNN_vs_Pretrained_Montage} shows that with the fusion technique, our proposed CNN achieves the best overall performance with an accuracy of 99.90\% and an AUC of 99.91\%. This demonstrates consistent results across all datasets, maintaining high precision, recall, F1-score, and AUC metrics for both Non-Responder (NR) and Responder (R) classes. It notably outperforms pretrained models including VGG16, Xception, ResNet152V2, MobileNetV2, and DenseNet201 in handling diverse time-frequency representations. It shows the strength of our custom CNN architecture in analyzing CWT, CWT\_ST montage, and CWT\_ST blending time-frequency images.
Comparing Tables~\ref{tab:pretrained_vs_cnn} and~\ref{tab:Performance_of_Proposed_CNN_vs_Pretrained_Montage}, we observe a significant improvement in accuracy when using fused time-frequency methods compared to individual time-frequency methods. The performance gains are evident across all models: VGG16 (from 83.94\% to 97.71\%), Xception (from 87.07\% to 94.40\%), ResNet152V2 (from 87.86\% to 96.55\%), MobileNetV2 (from 90.13\% to 98.84\%), DenseNet201 (from 90.78\% to 98.96\%), and the Proposed CNN (from 98.67\% to 99.90\%). In Fig.~\ref{beforeAfter}, the bar plot shows the accuracy comparison of individual time-frequency method (CWT) indicated as `Before' and fused time-frequency method (CWT\_ST) as `After'. This demonstrates that the fusion technique enhances the feature extraction capability for both pretrained models and the Proposed CNN. While individual time-frequency methods yield good predictions for MDD, combining two different time-frequency methods enables extraction of more discriminative features, achieving the highest prediction accuracy for classifying Responder and Non Responder.
Fig~\ref{fig:combined_CM_figures} shows the confusion matrix of VGG16, DenseNet201 and Proposed CNN.
\subsection{Testing with Secondary rTMS database}
To justify our model's feature extraction capability, we conduct a separate experiment using a different and significantly larger dataset, approximately three times the size of the primary one, which we refer to as the secondary dataset. In Table~\ref{tab:Performance_of_CNN_Independent}, we present a comparison of the four time-frequency methods and their corresponding accuracies using our custom CNN. The CWT method achieves the highest accuracy of 98.67\% among others. From our previous experiment shown in Table~\ref{tab:Performance_of_Fused_Methods}, we observe that fusion techniques montage and blending perform exceptionally well compared to individual time-frequency methods. To further validate this finding, we again apply these fusion techniques to combine two time-frequency methods and generate two distinct datasets. Using the montage fusion approach, we create CWT\_ST, CWT\_SSWE, and CWT\_DWT images. Similarly, using the blending technique, we generate CWT\_ST, CWT\_SSWE, and CWT\_DWT fused images. In Table~\ref{tab:Performance_of_Fused_Methods_testData}, we present a performance comparison of both montage and blending fusion techniques using our custom CNN. Consistent with the results from our primary dataset, the CWT\_ST and CWT\_SSWE combinations in both montage and blending configurations extract more relevant features and achieve higher accuracy in the secondary dataset. These findings reinforce that fusing multiple time-frequency methods leads to better classification performance.
\begin{table}[ht]
\centering
\resizebox{0.8\textwidth}{!}{%
\begin{tabular}{|c|cc|cc|cc|cc|c|c|}
\hline
\multirow{2}{*}{Fused} & \multicolumn{2}{c|}{Precision \%} & \multicolumn{2}{c|}{Recall \%} & \multicolumn{2}{c|}{Specificity \%} & \multicolumn{2}{c|}{$F_{\text{score}}$ \%} & \multirow{2}{*}{AUC} & \multirow{2}{*}{Accuracy \%} \\
\cline{2-9}
  & NR & R & NR & R & NR & R & NR & R &  & \\
\hline
\multicolumn{11}{|c|}{\textbf{Montage}} \\
\hline
CWT\_SSWE & 90 & 91 & 90 & 90 & 90 & 90 & 90 & 91 & 90.34 & 90.34 \\
CWT\_DWT & 89 & 88 & 89 & 89 & 89 & 88 & 88 & 89 & 88.59 & 88.61 \\
\textbf{CWT\_ST} & \textbf{92} & \textbf{92} & \textbf{91} & \textbf{92} & \textbf{92} & \textbf{92} & \textbf{92} & \textbf{92} & \textbf{91.89} & \textbf{91.90} \\
\hline
\multicolumn{11}{|c|}{\textbf{Blend}} \\
\hline
CWT\_SSWE & 89 & 90 & 90 & 89 & 89 & 90 & 89 & 90 & 89.59 & 89.59 \\
CWT\_DWT & 90 & 90 & 90 & 90 & 90 & 90 & 90 & 90 & 90.06 & 90.07 \\
\textbf{CWT\_ST} & \textbf{91} & \textbf{92} & \textbf{91} & \textbf{92} & \textbf{91} & \textbf{92} & \textbf{91} & \textbf{92} & \textbf{91.43} & \textbf{91.43} \\
\hline
\end{tabular}
}
\caption{Performance measures of Proposed CNN with fused methods on Secondary Dataset}\label{tab:Performance_of_Fused_Methods_testData}
\end{table}
To ensure our proposed CNN's feature extraction capability on the secondary dataset, we perform another set of experiments comparing our custom CNN with state-of-the-art models. This time we use CWT\_ST for both the montage and blending technique. Since DenseNet201 (98.96\%), MobileNetV2 (98.84\%) and VGG16 (97.71\%) are the top three accuracy achievers shown in Table~\ref{tab:Performance_of_Proposed_CNN_vs_Pretrained_Montage}, we compare the performance of these three architectures with our custom CNN.
\begin{table}[ht]
\centering
\resizebox{0.8\textwidth}{!}{%
\begin{tabular}{|c|cc|cc|cc|cc|c|c|}
\hline
\multirow{2}{*}{Fused} & \multicolumn{2}{c|}{Precision \%} & \multicolumn{2}{c|}{Recall \%} & \multicolumn{2}{c|}{Specificity \%} & \multicolumn{2}{c|}{$F_{\text{score}}$ \%} & \multirow{2}{*}{AUC} & \multirow{2}{*}{Accuracy \%} \\
\cline{2-9}
  & NR & R & NR & R & NR & R & NR & R &  & \\
\hline
\multicolumn{11}{|c|}{\textbf{Montage}} \\
\hline
VGG16 & 72 & 75 & 74 & 74 & 74 & 74 & 73 & 74 & 73.86 & 73.86 \\
ResNet152V2 & 74 & 76 & 76 & 75 & 76 & 75 & 75 & 76 & 75.43 & 75.42 \\
MobileNetV2 & 77 & 78 & 77 & 78 & 77 & 78 & 77 & 78 & 77.72 & 77.74 \\
\textbf{CNN} & \textbf{92} & \textbf{92} & \textbf{91} & \textbf{92} & \textbf{92} & \textbf{92} & \textbf{92} & \textbf{92} & \textbf{91.89} & \textbf{91.90} \\
\hline
\multicolumn{11}{|c|}{\textbf{Blend}} \\
\hline
VGG16 & 75 & 75 & 74 & 76 & 74 & 76 & 74 & 76 & 75.02 & 75.04 \\
ResNet152V2 & 75 & 77 & 76 & 76 & 76 & 76 & 75 & 76 & 75.71 & 75.71 \\
MobileNetV2 & 75 & 78 & 78 & 75 & 78 & 75 & 76 & 77 & 76.50 & 76.47 \\
\textbf{CNN} & \textbf{91} & \textbf{92} & \textbf{91} & \textbf{92} & \textbf{91} & \textbf{92} & \textbf{91} & \textbf{92} & \textbf{91.43} & \textbf{91.43} \\
\hline
\end{tabular}}
\caption{Performance measures of Proposed CNN and Pre-trained models on Secondary Dataset}\label{tab:Performance_of_CNN_and_pretrained_testData}
\end{table}
The comparative analysis shown in Table~\ref{tab:Performance_of_CNN_and_pretrained_testData} reveals that our CNN architecture consistently outperforms established pretrained models across both fusion techniques. In the montage approach, the proposed CNN achieves 91.90\% accuracy, with substantial improvements over VGG16 (73.86\%), ResNet152V2 (75.42\%), and MobileNetV2 (77.74\%). Similar performance advantages appear in the blend method, with our model attaining 91.43\% accuracy compared to 75.04--76.47\% for pretrained networks. The CNN maintains balanced performance across all metrics, with F1-scores of 92\% (NR) and 92\% (R) for montage fusion. This shows robust capability in classifying both responsive and non-responsive cases. The 14--18\% absolute accuracy gain over state-of-the-art pretrained models highlights the effectiveness of our custom architecture for this specific classification task, while the consistent results across different fusion techniques indicate method stability.
{\color{black}
\subsection{Subject-Level Cross-Validation Results}
\subsubsection{Subject-Disjoint Cross-Validation Protocol}
Unlike the segment-level setup in Section 3.4 where individual 20-second EEG segments are randomly partitioned across folds, this section applies a strict subject-disjoint protocol. All segments belonging to one patient stay in one fold only. This ensures no patient appears in both training and testing at any point, which removes the risk of patient-level data leakage.

We use stratified group three-fold cross-validation on the primary dataset. Each fold holds out five patients, three responders and two non-responders, and trains on the remaining ten.
We do not use leave-one-subject-out on this cohort. The reason is worth stating in full because it is easy to miss. The primary cohort has 9 responders and 6 non-responders. Holding out one patient leaves a training set of either 8 responders and 6 non-responders, or 9 responders and 5 non-responders. Only two training compositions are possible. Which one occurs is fixed by the label of the held-out patient. A model that is sensitive to the training class balance can therefore sort the held-out patients by their own labels. It does not need any information from the EEG signal to do this. With three folds of five patients every training set contains exactly 6 responders and 4 non-responders. The class balance is then identical in every fold and this route to leakage is closed.

On the secondary dataset we use stratified group five-fold cross-validation. That cohort has 23 responders and 23 non-responders, so each fold holds out about nine patients and the training class balance stays even from fold to fold. The prior leak described above does not arise there.

We pool the out-of-sample predictions from all folds and compute the metrics once over the complete cohort. Every patient contributes exactly one prediction, and that prediction comes from a model that never saw that patient. Confidence intervals come from a bootstrap that resamples patients, which is the correct unit of analysis.

Three further points apply throughout. Early stopping and the learning rate schedule watch an inner validation split drawn only from the training patients, so the held-out patients are never used during training. We apply balanced class weights, because the primary cohort has more responder than non-responder segments. We select the decision threshold on the inner validation split using Youden's J rather than fixing it at 0.5. The whole procedure is repeated with five random initialization seeds, and we report the mean and standard deviation across seeds.

\subsubsection{Testing with Primary rTMS database}
Table \ref{tab:sub_sm_gen} shows the subject-level results for the four standalone TF methods on the primary dataset. No method separates responders from non-responders. ST reaches the highest AUC at $0.485 \pm 0.079$ and CWT follows at $0.478 \pm 0.072$, with DWT and SSWE below both. Every AUC 95\% confidence interval contains 0.5 and every interval is wide, spanning roughly 0.14 to 0.81, which is what 15 patients yields when there is nothing to detect. Sensitivity and specificity trade off against each other in every row, so a method that appears sensitive is simply operating at a permissive threshold rather than detecting anything.
\begin{table}[H]
\scriptsize
  \centering
  \caption{Subject-level performance of the four standalone time-frequency methods on the primary dataset, under stratified group three-fold cross-validation with five random seeds. All four sit at chance. The base-rate Brier reference on this cohort is 0.24.}
  \label{tab:sub_sm_gen}
  \begin{tabular}{|c|c|c|c|c|c|c|c|}
    \hline
    \textbf{Method} &
    \textbf{AUC} & \textbf{AUC 95\% CI} &
    \textbf{Sensitivity} & \textbf{Specificity} &
    \textbf{F1-Score} &
    \textbf{Accuracy} & \textbf{Brier}
     \\
    \hline
    ST & 0.485 ± 0.079 & 0.170--0.804 & 0.533 ± 0.183 & 0.467 ± 0.217 & 0.552 ± 0.097 & 0.507 ± 0.037 & 0.287 \\
    \hline
    CWT & 0.478 ± 0.072 & 0.138--0.817 & 0.711 ± 0.169 & 0.300 ± 0.183 & 0.647 ± 0.061 & 0.547 ± 0.030 & 0.276 \\
    \hline
    DWT & 0.452 ± 0.085 & 0.135--0.785 & 0.578 ± 0.145 & 0.367 ± 0.183 & 0.569 ± 0.085 & 0.493 ± 0.037 & 0.288 \\
    \hline
    SSWE & 0.441 ± 0.042 & 0.137--0.769 & 0.444 ± 0.208 & 0.567 ± 0.224 & 0.491 ± 0.134 & 0.493 ± 0.037 & 0.292 \\
    \hline
  \end{tabular}
\end{table}

We pair CWT with DWT, SSWE, and ST using both the montage and blend strategies. Table \ref{tab:dataset_modalities_grouped} shows the results. Five of the six pairings sit at chance, between $0.444$ and $0.481$, with every confidence interval containing 0.5. Blend CWT\_ST reaches $0.474 \pm 0.098$ against $0.478$ for CWT alone and $0.485$ for ST alone, so combining the two representations recovers less than either one on its own.
The best result is montage CWT\_ST, which reaches $0.874 \pm 0.183$ with a confidence interval of 0.718 to 0.975 and 82.7\% accuracy. This is the highest subject-level performance obtained anywhere in this study.

Three observations bound how far this result can be generalised at present. Its two components, CWT and ST, are at chance individually. Blend CWT\_ST is at chance, and blend uses the identical two source images, differing only in whether they are placed side by side or overlaid. The same montage strategy reaches 0.481 and 0.474 on the CWT\_DWT and CWT\_SSWE pairings, and 0.457 to 0.514 on all three pairings of the secondary cohort. Montage keeps both representations intact side by side. Blend averages the two images into one and loses part of each. This is consistent with montage outperforming blend on the same pair of methods.

\begin{table}[H]
\scriptsize
  \centering
  \caption{Subject-level performance of montage and blend fusion on the primary dataset, under stratified group three-fold cross-validation with five random seeds. Montage CWT\_ST gives the best result in this study.}
  \label{tab:dataset_modalities_grouped}
  \begin{tabular}{|c|c|c|c|c|c|c|c|c|}
    \hline
    \textbf{Strategy} & \textbf{Method} & \textbf{AUC} & \textbf{AUC 95\% CI} & \textbf{Sensitivity} & \textbf{Specificity} & \textbf{F1-Score} & \textbf{Accuracy} & \textbf{Brier} \\
    \hline
    \multirow{3}{*}{Blend}
    & CWT\_DWT  & 0.478 ± 0.063 & 0.147--0.810 & 0.556 ± 0.157 & 0.433 ± 0.149 & 0.565 ± 0.095 & 0.507 ± 0.037 & 0.289 \\
    \cline{2-9}
    & CWT\_ST   & 0.474 ± 0.098 & 0.147--0.808 & 0.667 ± 0.208 & 0.333 ± 0.204 & 0.617 ± 0.112 & 0.533 ± 0.047 & 0.300 \\
    \cline{2-9}
    & CWT\_SSWE & 0.444 ± 0.113 & 0.139--0.769 & 0.533 ± 0.298 & 0.400 ± 0.303 & 0.521 ± 0.150 & 0.480 ± 0.073 & 0.299 \\
    \hline
    \hline
    \multirow{3}{*}{Montage}
    & CWT\_DWT  & 0.481 ± 0.090 & 0.158--0.825 & 0.533 ± 0.145 & 0.500 ± 0.167 & 0.563 ± 0.094 & 0.520 ± 0.056 & 0.294 \\
    \cline{2-9}
    & CWT\_SSWE & 0.474 ± 0.070 & 0.164--0.792 & 0.400 ± 0.099 & 0.567 ± 0.149 & 0.468 ± 0.058 & 0.467 ± 0.000 & 0.284 \\
    \cline{2-9}
    & CWT\_ST$^{\dagger}$ & 0.874 ± 0.183 & 0.718--0.975 & 0.933 ± 0.149 & 0.667 ± 0.236 & 0.866 ± 0.138 & 0.827 ± 0.174 & \textbf{0.130} \\
    \hline
  \end{tabular}
\end{table}

\begin{sloppypar} \textcolor{black}{We compare the proposed CNN against thirteen alternatives spanning four families: five pretrained computer vision models (VGG16, Xception, ResNet152V2, MobileNetV2, DenseNet201), three EEG-specific architectures (EEGNet, EEGNetV2, SleepEEGNet), a CNN-LSTM hybrid, and four classical machine learning classifiers on patient-level summary features (logistic regression, linear SVM, RBF SVM, random forest). All use the identical folds, seeds, class weights, threshold rule and per-patient pooling as the proposed model, so every number in Tables~\ref{tab:sota_primary_cwt}, \ref{tab:sota_primary_montage} and~\ref{tab:sota_secondary} is directly comparable. We run all fourteen models on two inputs: the standalone CWT images and the montage CWT\_ST images.}\end{sloppypar}

\textcolor{black}{Two adaptations should be stated. The EEG-specific architectures are designed for raw EEG arranged as channels by time. We convert each image to grayscale and treat the vertical axis as the channel dimension and the horizontal axis as time, which is the standard adaptation for spectrogram input. They are not run on raw EEG. The classical classifiers operate on 64 features per patient: the image intensity profile along the vertical axis reduced to 32 bins, plus the standard deviation of each bin across that patient's images. This is the image-domain analogue of a band-power summary, not a band-power analysis of the original signal.}

\textcolor{black}{The input decides the result and the model barely matters. On the standalone CWT images all fourteen sit at chance, from 0.307 to 0.541. The proposed CNN at 0.478 falls in the middle of that range. On the montage CWT\_ST images six of the fourteen reach 0.82 or above. The proposed CNN reaches 0.874 and ranks fifth. SleepEEGNet and MobileNetV2 both reach 0.944. SleepEEGNet beats the proposed model on every metric, with 90.7\% accuracy, 0.867 specificity and a Brier score of 0.060 against our 0.130. We state this plainly rather than claim an advantage the data do not support.}

\textcolor{black}{Table~\ref{tab:model_size} sets these results against model size. The proposed CNN is the smallest of the six architectures that exceed 0.82 on this input. SleepEEGNet has 1.9 times as many parameters, CNN-LSTM 2.4 times, MobileNetV2 6.4 times, VGG16 36 times and DenseNet201 46 times. Our model reaches the same 82.7\% accuracy as CNN-LSTM and DenseNet201 with a fraction of their capacity, which is the trade the lightweight design was intended to make.}

\begin{table}[H]
\scriptsize
  \centering
  \caption{\textcolor{black}{Model size for the six architectures that exceed AUC 0.82 on the montage CWT\_ST input of the primary cohort. The convolutional base of the pretrained models is frozen, so the trainable count is small while the total count sets memory use and inference cost.}}
  \label{tab:model_size}
  \begin{tabular}{|l|l|c|c|r|r|}
    \hline
    \textbf{Model} & \textbf{Family} & \textbf{AUC} & \textbf{Accuracy} & \textbf{Total parameters} & \textbf{Trainable} \\
    \hline
    SleepEEGNet & EEG-specific & 0.944 & 0.907 & 771,585 & 771,329 \\
    \hline
    MobileNetV2 & Pretrained & 0.944 & 0.813 & 2,591,297 & 330,753 \\
    \hline
    CNN-LSTM & Hybrid & 0.926 & 0.827 & 974,945 & 974,721 \\
    \hline
    DenseNet201 & Pretrained & 0.922 & 0.827 & 18,821,697 & 495,873 \\
    \hline
    \textbf{Proposed CNN} & \textbf{Proposed} & \textbf{0.874} & \textbf{0.827} & \textbf{407,665} & \textbf{407,665} \\
    \hline
    VGG16 & Pretrained & 0.826 & 0.773 & 14,848,321 & 132,609 \\
    \hline
  \end{tabular}
\end{table}

\textcolor{black}{Table~\ref{tab:sota_secondary} repeats the whole comparison on the secondary cohort. All fourteen models span 0.270 to 0.630 on the standalone ST images and 0.313 to 0.621 on the montage ST\_CWT images. No model on either input exceeds 0.63 and every confidence interval contains 0.5. Seven of the fourteen score higher on the montage input than on the standalone one and seven score lower.}

\begin{table}[H]
\scriptsize
  \centering
  \caption{Fourteen models on the primary cohort, standalone CWT input, under stratified group three-fold cross-validation with five seeds. Sorted by AUC. Every model sits at chance.}
  \label{tab:sota_primary_cwt}
  \begin{tabular}{|l|l|c|c|c|c|c|}
    \hline
    \textbf{Model} & \textbf{Family} & \textbf{AUC} & \textbf{Sensitivity} & \textbf{Specificity} & \textbf{F1-Score} & \textbf{Accuracy} \\
    \hline
    SVM-RBF & Classical & 0.541 ± 0.079 & 0.578 & 0.667 & 0.639 & 0.613 \\
    \hline
    ResNet152V2 & Pretrained & 0.537 ± 0.112 & 0.778 & 0.233 & 0.660 & 0.560 \\
    \hline
    Logistic regression & Classical & 0.526 ± 0.066 & 0.733 & 0.533 & 0.704 & 0.653 \\
    \hline
    Random forest & Classical & 0.507 ± 0.098 & 0.600 & 0.667 & 0.604 & 0.627 \\
    \hline
    CNN-LSTM & Hybrid & 0.496 ± 0.076 & 0.556 & 0.433 & 0.545 & 0.507 \\
    \hline
    Proposed CNN & Proposed & 0.478 ± 0.072 & 0.711 & 0.300 & 0.647 & 0.547 \\
    \hline
    EEGNet & EEG-specific & 0.448 ± 0.085 & 0.400 & 0.500 & 0.411 & 0.440 \\
    \hline
    DenseNet201 & Pretrained & 0.441 ± 0.136 & 0.600 & 0.400 & 0.564 & 0.520 \\
    \hline
    MobileNetV2 & Pretrained & 0.433 ± 0.126 & 0.533 & 0.467 & 0.540 & 0.507 \\
    \hline
    EEGNetV2 & EEG-specific & 0.430 ± 0.076 & 0.556 & 0.400 & 0.552 & 0.493 \\
    \hline
    Xception & Pretrained & 0.411 ± 0.109 & 0.356 & 0.467 & 0.331 & 0.400 \\
    \hline
    SleepEEGNet & EEG-specific & 0.404 ± 0.091 & 0.489 & 0.500 & 0.482 & 0.493 \\
    \hline
    VGG16 & Pretrained & 0.370 ± 0.168 & 0.733 & 0.300 & 0.649 & 0.560 \\
    \hline
    SVM-linear & Classical & 0.307 ± 0.048 & 0.689 & 0.333 & 0.598 & 0.547 \\
    \hline
  \end{tabular}
\end{table}

\begin{table}[H]
\scriptsize
  \centering
  \caption{The same fourteen models on the primary cohort, montage CWT\_ST input, under the identical protocol. Six exceed 0.82. The highest values belong to SleepEEGNet and MobileNetV2 rather than to the proposed CNN.}
  \label{tab:sota_primary_montage}
  \begin{tabular}{|l|l|c|c|c|c|c|}
    \hline
    \textbf{Model} & \textbf{Family} & \textbf{AUC} & \textbf{Sensitivity} & \textbf{Specificity} & \textbf{F1-Score} & \textbf{Accuracy} \\
    \hline
    \textbf{SleepEEGNet} & EEG-specific & \textbf{0.944 ± 0.114} & \textbf{0.933} & \textbf{0.867} & \textbf{0.921} & \textbf{0.907} \\
    \hline
    MobileNetV2 & Pretrained & 0.944 ± 0.069 & 0.867 & 0.733 & 0.842 & 0.813 \\
    \hline
    CNN-LSTM & Hybrid & 0.926 ± 0.086 & 1.000 & 0.567 & 0.877 & 0.827 \\
    \hline
    DenseNet201 & Pretrained & 0.922 ± 0.082 & 0.822 & 0.833 & 0.841 & 0.827 \\
    \hline
    Proposed CNN & Proposed & 0.874 ± 0.183 & 0.933 & 0.667 & 0.866 & 0.827 \\
    \hline
    VGG16 & Pretrained & 0.826 ± 0.152 & 0.778 & 0.767 & 0.802 & 0.773 \\
    \hline
    Logistic regression & Classical & 0.722 ± 0.107 & 0.844 & 0.700 & 0.820 & 0.787 \\
    \hline
    Xception & Pretrained & 0.656 ± 0.137 & 0.667 & 0.500 & 0.643 & 0.600 \\
    \hline
    ResNet152V2 & Pretrained & 0.619 ± 0.171 & 0.644 & 0.400 & 0.587 & 0.547 \\
    \hline
    Random forest & Classical & 0.611 ± 0.063 & 0.844 & 0.533 & 0.776 & 0.720 \\
    \hline
    EEGNetV2 & EEG-specific & 0.563 ± 0.091 & 0.600 & 0.433 & 0.587 & 0.533 \\
    \hline
    EEGNet & EEG-specific & 0.474 ± 0.084 & 0.600 & 0.333 & 0.561 & 0.493 \\
    \hline
    SVM-linear & Classical & 0.363 ± 0.211 & 0.867 & 0.300 & 0.744 & 0.640 \\
    \hline
    SVM-RBF & Classical & 0.333 ± 0.094 & 0.578 & 0.500 & 0.535 & 0.547 \\
    \hline
  \end{tabular}
\end{table}

\begin{table}[H]
\scriptsize
  \centering
  \caption{The same fourteen models on the secondary cohort, both inputs, under stratified group five-fold cross-validation with five seeds. No model on either input exceeds 0.63.}
  \label{tab:sota_secondary}
  \begin{tabular}{|l|l|c|c|c|c|c|}
    \hline
    \textbf{Model} & \textbf{Family} & \textbf{AUC} & \textbf{Sensitivity} & \textbf{Specificity} & \textbf{F1-Score} & \textbf{Accuracy} \\
    \hline
    \multicolumn{7}{|c|}{\textbf{Input: standalone ST}} \\
    \hline
    SVM-linear & Classical & 0.630 ± 0.081 & 0.609 & 0.722 & 0.620 & 0.665 \\
    \hline
    SVM-RBF & Classical & 0.556 ± 0.069 & 0.722 & 0.513 & 0.641 & 0.617 \\
    \hline
    VGG16 & Pretrained & 0.491 ± 0.091 & 0.426 & 0.565 & 0.407 & 0.496 \\
    \hline
    Random forest & Classical & 0.483 ± 0.047 & 0.817 & 0.348 & 0.662 & 0.583 \\
    \hline
    Proposed CNN & Proposed & 0.470 ± 0.030 & 0.504 & 0.435 & 0.474 & 0.470 \\
    \hline
    CNN-LSTM & Hybrid & 0.464 ± 0.035 & 0.617 & 0.365 & 0.541 & 0.491 \\
    \hline
    EEGNetV2 & EEG-specific & 0.455 ± 0.027 & 0.557 & 0.417 & 0.495 & 0.487 \\
    \hline
    ResNet152V2 & Pretrained & 0.447 ± 0.107 & 0.400 & 0.522 & 0.349 & 0.461 \\
    \hline
    DenseNet201 & Pretrained & 0.441 ± 0.059 & 0.478 & 0.487 & 0.432 & 0.483 \\
    \hline
    EEGNet & EEG-specific & 0.439 ± 0.065 & 0.548 & 0.365 & 0.483 & 0.457 \\
    \hline
    SleepEEGNet & EEG-specific & 0.439 ± 0.055 & 0.530 & 0.409 & 0.432 & 0.470 \\
    \hline
    MobileNetV2 & Pretrained & 0.425 ± 0.060 & 0.461 & 0.487 & 0.411 & 0.474 \\
    \hline
    Xception & Pretrained & 0.386 ± 0.113 & 0.504 & 0.435 & 0.411 & 0.470 \\
    \hline
    Logistic regression & Classical & 0.270 ± 0.055 & 0.800 & 0.209 & 0.535 & 0.504 \\
    \hline
    \multicolumn{7}{|c|}{\textbf{Input: montage ST\_CWT}} \\
    \hline
    SVM-linear & Classical & 0.621 ± 0.017 & 0.774 & 0.496 & 0.676 & 0.635 \\
    \hline
    SVM-RBF & Classical & 0.608 ± 0.076 & 0.678 & 0.609 & 0.645 & 0.643 \\
    \hline
    Proposed CNN & Proposed & 0.514 ± 0.068 & 0.617 & 0.409 & 0.529 & 0.513 \\
    \hline
    DenseNet201 & Pretrained & 0.489 ± 0.125 & 0.487 & 0.496 & 0.410 & 0.491 \\
    \hline
    MobileNetV2 & Pretrained & 0.488 ± 0.083 & 0.643 & 0.357 & 0.544 & 0.500 \\
    \hline
    ResNet152V2 & Pretrained & 0.476 ± 0.094 & 0.748 & 0.226 & 0.591 & 0.487 \\
    \hline
    Random forest & Classical & 0.449 ± 0.027 & 0.809 & 0.278 & 0.635 & 0.543 \\
    \hline
    CNN-LSTM & Hybrid & 0.444 ± 0.045 & 0.652 & 0.365 & 0.558 & 0.509 \\
    \hline
    EEGNetV2 & EEG-specific & 0.431 ± 0.093 & 0.435 & 0.504 & 0.373 & 0.470 \\
    \hline
    VGG16 & Pretrained & 0.421 ± 0.129 & 0.383 & 0.557 & 0.335 & 0.470 \\
    \hline
    SleepEEGNet & EEG-specific & 0.399 ± 0.096 & 0.496 & 0.391 & 0.459 & 0.443 \\
    \hline
    EEGNet & EEG-specific & 0.396 ± 0.084 & 0.635 & 0.278 & 0.527 & 0.457 \\
    \hline
    Xception & Pretrained & 0.389 ± 0.121 & 0.704 & 0.261 & 0.518 & 0.483 \\
    \hline
    Logistic regression & Classical & 0.313 ± 0.028 & 0.991 & 0.017 & 0.667 & 0.504 \\
    \hline
  \end{tabular}
\end{table}

Table \ref{tab:dataset2_modalities} shows the subject-level results for the four standalone TF methods on the secondary dataset. As on the primary dataset, no method separates responders from non-responders. CWT and DWT both reach $0.495$, SSWE $0.479$ and ST $0.470$. Every AUC 95\% confidence interval contains 0.5. The intervals are tighter here than on the primary dataset, spanning roughly 0.30 to 0.67 rather than 0.14 to 0.81, because 46 patients estimate the quantity more precisely than 15. That tighter estimate is centred on chance.

This cohort has three times as many patients and balanced classes, so its estimates are more precise. The two cohorts are not interchangeable, however; they were recorded at different times, and their fused images were generated with a different anchor method, CWT on the primary cohort and ST here, so the montage inputs are not the same set of images.

\begin{table}[H]
\scriptsize
  \centering
  \caption{Subject-level performance of the four standalone time-frequency methods on the secondary dataset, under stratified group five-fold cross-validation with five random seeds. All four sit at chance.}
  \label{tab:dataset2_modalities}
  \begin{tabular}{|c|c|c|c|c|c|c|c|}
    \hline
    \textbf{Method} &
    \textbf{AUC} & \textbf{AUC 95\% CI} &
    \textbf{Sensitivity} & \textbf{Specificity} &
    \textbf{F1-Score} &
    \textbf{Accuracy} & \textbf{Brier}
     \\
    \hline
    CWT & 0.495 ± 0.033 & 0.320--0.663 & 0.652 ± 0.182 & 0.330 ± 0.136 & 0.554 ± 0.074 & 0.491 ± 0.033 & 0.333 \\
    \hline
    DWT & 0.495 ± 0.017 & 0.321--0.666 & 0.478 ± 0.187 & 0.530 ± 0.200 & 0.478 ± 0.086 & 0.504 ± 0.018 & 0.328 \\
    \hline
    SSWE & 0.479 ± 0.030 & 0.308--0.651 & 0.470 ± 0.094 & 0.530 ± 0.094 & 0.481 ± 0.048 & 0.500 ± 0.000 & 0.347 \\
    \hline
    ST & 0.470 ± 0.030 & 0.299--0.643 & 0.504 ± 0.201 & 0.435 ± 0.215 & 0.474 ± 0.088 & 0.470 ± 0.012 & 0.347 \\
    \hline
  \end{tabular}
\end{table}

We pair ST with DWT, SSWE, and CWT using both the montage and blend strategies. Table \ref{tab:dataset2_modalities_grouped} shows the results. The six pairings fall between $0.457$ and $0.529$, and every confidence interval contains 0.5. On this cohort the fused inputs track their standalone components, which span $0.470$ to $0.495$.

Montage ST\_CWT reaches $0.514 \pm 0.068$ on this cohort, against $0.874$ for montage CWT\_ST on the primary cohort in Table \ref{tab:dataset_modalities_grouped}. The two are not the same input. The cohorts were recorded at different times, and their fused images were generated with a different anchor method, CWT on the primary cohort and ST here, so the two halves of the montage are in the opposite order.

These results sit far below the segment-level results in Section 4.4. The gap is the effect of removing patient-level leakage, and measuring it is one of the contributions of this study. Under the segment-level protocol, segments from the same patient appear in both training and testing, so the network can identify the patient and recover the label from that. Under the subject-disjoint protocol no patient appears in both, and that route is closed. This study reports both protocols on both cohorts. None of the studies in Table \ref{tab:comparison_rTMS_methods} states how its folds were built, and none reports a subject-independent result, so this is the only evaluation in this area that can be read without that ambiguity.

\begin{table}[H]
\scriptsize
  \centering
  \caption{Subject-level performance of montage and blend fusion on the secondary dataset, under stratified group five-fold cross-validation with five random seeds. No pairing exceeds chance under either strategy.}
  \label{tab:dataset2_modalities_grouped}
  \begin{tabular}{|c|c|c|c|c|c|c|c|c|}
    \hline
    \textbf{Strategy} & \textbf{Method} & \textbf{AUC} & \textbf{AUC 95\% CI} & \textbf{Sensitivity} & \textbf{Specificity} & \textbf{F1-Score} & \textbf{Accuracy} & \textbf{Brier} \\
    \hline
    \multirow{3}{*}{Blend}
    & ST\_CWT  & 0.529 ± 0.072 & 0.360--0.696 & 0.574 ± 0.235 & 0.426 ± 0.267 & 0.516 ± 0.122 & 0.500 ± 0.074 & 0.334 \\
    \cline{2-9}
    & ST\_DWT  & 0.505 ± 0.093 & 0.338--0.673 & 0.678 ± 0.245 & 0.278 ± 0.205 & 0.546 ± 0.131 & 0.478 ± 0.027 & 0.332 \\
    \cline{2-9}
    & ST\_SSWE & 0.488 ± 0.064 & 0.321--0.657 & 0.696 ± 0.174 & 0.296 ± 0.223 & 0.573 ± 0.058 & 0.496 ± 0.036 & 0.344 \\
    \hline
    \hline
    \multirow{3}{*}{Montage}
    & ST\_CWT  & 0.514 ± 0.068 & 0.342--0.686 & 0.617 ± 0.286 & 0.409 ± 0.191 & 0.529 ± 0.176 & 0.513 ± 0.059 & 0.351 \\
    \cline{2-9}
    & ST\_SSWE & 0.500 ± 0.089 & 0.331--0.669 & 0.530 ± 0.185 & 0.487 ± 0.200 & 0.505 ± 0.100 & 0.509 ± 0.033 & 0.348 \\
    \cline{2-9}
    & ST\_DWT  & 0.457 ± 0.094 & 0.290--0.622 & 0.652 ± 0.378 & 0.339 ± 0.329 & 0.496 ± 0.282 & 0.496 ± 0.054 & 0.342 \\
    \hline
  \end{tabular}
\end{table}
\subsubsection{Summary of Significance Testing}
We test the primary-cohort results with a patient-level permutation test run under the same three-fold protocol. We shuffle the responder and non-responder labels across whole patients and rerun the complete cross-validation pipeline. Table~\ref{tab:significance} reports the test for the best standalone method, ST, and for the montage CWT\_ST configuration.

For ST the observed AUC of 0.500 falls in the middle of the null distribution, which has a mean of 0.520 and a standard deviation of 0.083. Seventy of the 100 permutations reach or exceed the observed value, giving $p = 0.703$. This is the expected outcome for a method operating at chance, and it applies equally to the other standalone and fused configurations, all of which fall in the same range.

For montage CWT\_ST the observed value exceeds every one of 200 permutations, giving $p = 0.005$. We evaluate eight fusion configurations on this cohort. Testing eight at once raises the chance that one looks significant by luck, so we tighten the threshold from 0.05 to 0.05 divided by 8, which is 0.00625. The observed $p$ of 0.005 is below that, so the result holds under the strict correction. We state the number of configurations explicitly so that a reader who prefers a different correction can apply one. We note that this configuration behaves differently from every other configuration in this study, including blend CWT\_ST which uses the identical two source images, and that the reason for the difference is not established.

A two-sample $t$-test on fold-level AUC values would not be appropriate here. Three folds give only three values per method, which is too few to test, and those values are not independent because every fold shares training patients with the other two. The permutation test has neither problem. It builds the null distribution by rerunning the complete pipeline on shuffled labels, so it needs no assumption about how the fold values are distributed.

\begin{table}[H]
\scriptsize
  \centering
  \caption{Patient-level permutation tests on the primary cohort. Labels are shuffled across whole patients and the complete cross-validation pipeline is rerun under the same stratified group three-fold protocol.}
  \label{tab:significance}
  \begin{tabular}{|l|c|c|}
    \hline
    \textbf{Quantity} & \textbf{ST} & \textbf{Montage CWT\_ST} \\
    \hline
    Permutations & 100 & 200 \\
    \hline
    Null mean AUC & 0.520 & 0.520 \\
    \hline
    Null standard deviation & 0.083 & 0.105 \\
    \hline
    Null 95th percentile & 0.630 & 0.704 \\
    \hline
    Null maximum & 0.759 & 0.815 \\
    \hline
    Observed AUC, reference seed & 0.500 & 1.000 \\
    \hline
    Observed AUC, mean over five seeds & 0.485 & 0.874 \\
    \hline
    Permutations reaching the observed value & 70 & 0 \\
    \hline
    $p$ & 0.703 & \textbf{0.005} \\
    \hline
    Bonferroni threshold, eight configurations & 0.00625 & 0.00625 \\
    \hline
  \end{tabular}
\end{table}
}

\begin{figure}[H]
\centering
\includegraphics[width=\textwidth]{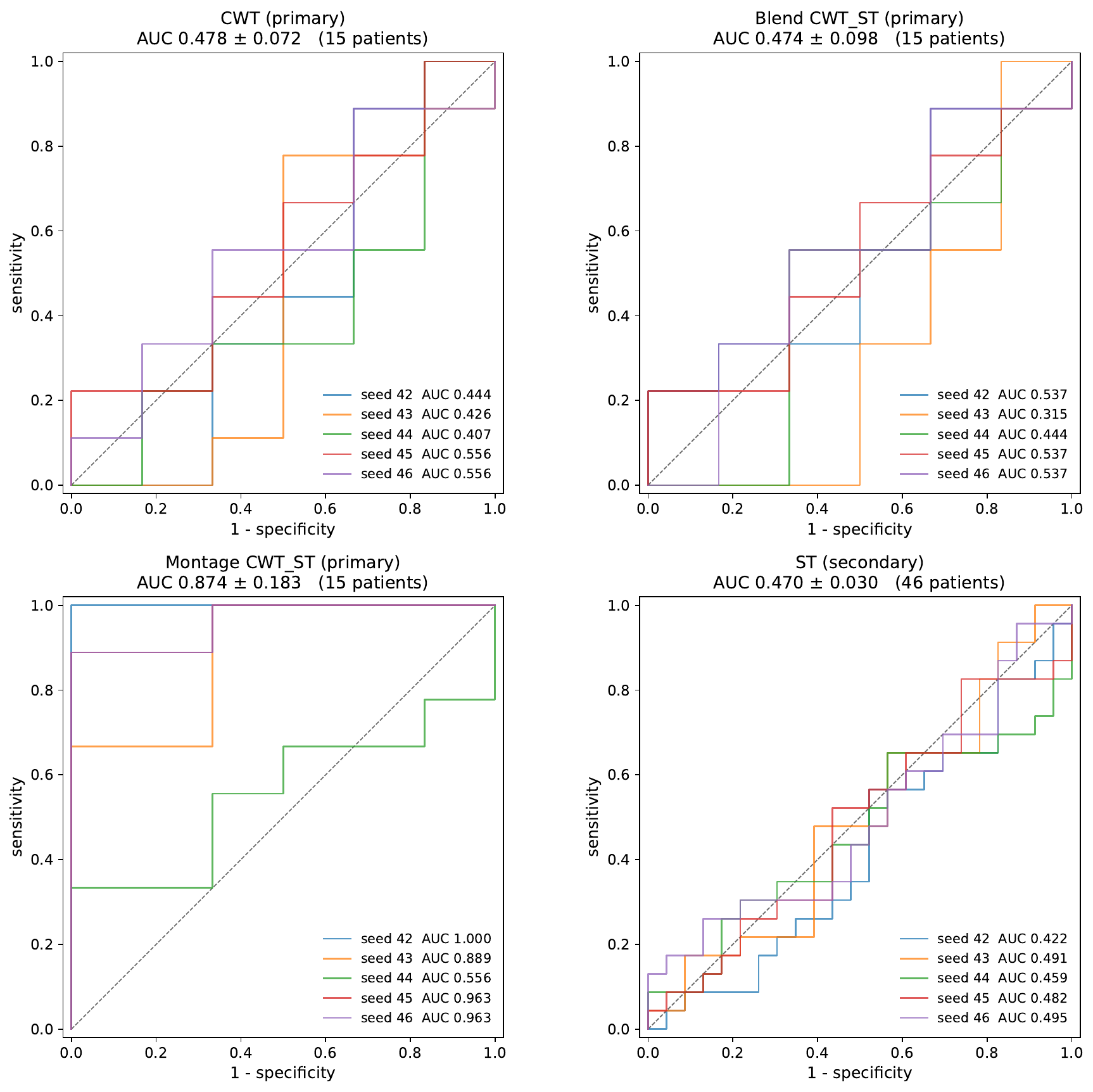}
\caption{\textcolor{black}{Subject-level ROC curves, one line per seed. The first three panels are the primary cohort under stratified group three-fold cross-validation. The fourth is the secondary cohort under five-fold. Each curve pools the out-of-sample predictions across folds, so every patient contributes one point.}}
\label{fig:roc_subject}
\end{figure}

\textcolor{black}{Figure~\ref{fig:roc_subject} shows the subject-level ROC curves for the two standalone methods and the two CWT\_ST fusions, with one line per seed.}

\section{Discussion}
In this study, we classify EEG signals to distinguish between responders and non-responders to rTMS therapy using a fused TF approach paired with a custom-designed CNN framework. We explore four distinct TF techniques: CWT, DWT, ST, and SSWE. To capture richer multi-modal signal characteristics, we introduce two image-level fusion strategies: blending and montage. \textcolor{black}{Under segment-level validation the fused TF representations outperform the standalone TF methods in every experiment. Under subject-disjoint validation one configuration does so, montage CWT\_ST on the primary cohort, which gives the best result in this study at AUC $0.874$ and 82.7\% accuracy.}

\subsection{Ablation Study - Proposed CNN and Pre-trained CNNs}
{\color{black}
Our proposed CNN has three convolutional layers with 8, 16, and 32 filters, each with a 3$\times$3 kernel and ReLU activation. We apply Batch Normalization after each convolutional layer to stabilize training. Average Pooling with a 2$\times$2 window reduces the spatial size of the feature maps. We then flatten the output and apply a Dropout layer with a rate of 0.25 to reduce overfitting. The final output is a single sigmoid node for binary classification. We train with the Adam optimizer at a learning rate of 0.0001 and Binary Cross-Entropy loss, and use early stopping with patience 5.

We apply the same training setup to all baseline models. For the pretrained vision models (VGG16, Xception, ResNet152V2, MobileNetV2, and DenseNet201), we initialize the convolutional base with ImageNet weights and keep it frozen. We attach a classification head consisting of Global Average Pooling, Batch Normalization, a Dense layer with 256 neurons and ReLU activation, a Dropout layer (rate 0.3 to 0.5 depending on model capacity), and a single sigmoid output. We apply the same setup to the EEG-specific models (EEGNet, EEGNetV2, and SleepEEGNet). All models use the Adam optimizer at 0.0001 with early stopping. Using the same configuration for all models ensures a fair comparison.
}
\subsection{Proposed CNN Architecture Exploration}
{\color{black}
To select the CNN architecture, we ran eight ablation experiments varying network depth, filter sizes, dropout rate, and pooling strategy. Exp-1 to Exp-3 test progressively deeper networks with more filters. Exp-4 and Exp-5 test deep architectures with large filter banks (64, 128, and 256 filters) and compare Average Pooling against Global Average Pooling. Exp-6 and Exp-7 test medium-depth networks with the same filter sizes but different dropout rates to isolate the effect of regularization.

This selection was carried out under the segment-level protocol, before the subject-disjoint analysis reported in Section 4.5. Under that protocol Exp-1 to Exp-3 achieve between 94.59\% and 95.24\%, Exp-4 and Exp-5 reach 99.86\% and 99.66\% with substantially more parameters, and Exp-8, the lightweight configuration with 8, 16, and 32 filters, Average Pooling, and a single dropout rate of 0.25, reaches 99.90\%. Exp-8 was selected on that basis and is used in every experiment in this paper.

We retain that choice rather than revisiting it, and we note two cases. The margin separating Exp-8 from Exp-4 is 0.04 percentage points, which is not a meaningful difference. And the protocol under which the comparison was made is the one Section 4.5 shows to be unreliable.

Table~\ref{tab:cnn_ablation_study} reports all eight architectures under both protocols. Both runs use the same code and the same training settings, and the only difference between the two sets of columns is how the folds are built. This isolates the inflation that segment-level validation produces. Under the segment-level protocol every one of the eight reaches a patient-level AUC of exactly 1.000. Under the subject-disjoint protocol every one returns a value between 0.431 and 0.493. The parameter count spans three orders of magnitude, from 389,121 to 205,545,057, and it changes neither figure, so the inflation belongs to the fold construction and not to any architecture.

This also bears on the architecture selection above. Under the segment-level protocol all eight configurations reach the same ceiling, so the 0.04 percentage point margin separating Exp-8 from Exp-4 is within run-to-run variation rather than a real difference. Under the subject-disjoint protocol the eight again fall in a narrow band. The choice among these eight architectures therefore has little effect on the reported figure, while the fold construction has a large one.
}
\begin{table}[h]
\fontsize{9pt}{11pt}\selectfont
\centering
\resizebox{\textwidth}{!}{
\begin{tabular}{|c|c|c|c|r|c|c|c|c|}
\hline
\multirow{2}{*}{\textbf{Experiment}} &
\multirow{2}{*}{\textbf{Purpose / Configuration}} &
\multirow{2}{*}{\textbf{Filters}} &
\multirow{2}{*}{\textbf{Pooling}} &
\multirow{2}{*}{\textbf{Parameters}} &
\multicolumn{2}{c|}{\textbf{Segment-level (leaky)}} &
\multicolumn{2}{c|}{\textbf{Subject-disjoint}} \\
\cline{6-9}
 & & & & & \textbf{Seg AUC} & \textbf{Patient AUC} & \textbf{Seg AUC} & \textbf{Patient AUC} \\
\hline
Exp-1 & Baseline shallow CNN & 4,6,8,12 & None & 603,883 & 0.995 & \textbf{1.000} & 0.443 & \textbf{0.437} \\
\hline
Exp-2 & Increased depth, additional 5x5 kernel & 4,6,8,12,32 & None & 1,617,163 & 0.995 & \textbf{1.000} & 0.474 & \textbf{0.437} \\
\hline
Exp-3 & Deepened CNN, larger filter bank & 4,6,8,12,32,64 & None & 3,274,315 & 0.992 & \textbf{1.000} & 0.464 & \textbf{0.459} \\
\hline
Exp-4 & Deep CNN with Average Pooling & 64,128,256 & AvgPool & 205,893,633 & 0.951 & \textbf{1.000} & 0.481 & \textbf{0.472} \\
\hline
Exp-5 & Deep CNN with Global Average Pooling & 64,128,256 & GlobalAvg & 389,121 & 0.861 & \textbf{1.000} & 0.456 & \textbf{0.461} \\
\hline
Exp-6 & Medium depth, standard dropout & 16,32,64 & None & 205,545,057 & 0.971 & \textbf{1.000} & 0.455 & \textbf{0.431} \\
\hline
Exp-7 & Medium depth, increased dropout & 16,32,64 & None & 205,545,057 & 0.982 & \textbf{1.000} & 0.470 & \textbf{0.493} \\
\hline
Exp-8 (Proposed) & Lightweight CNN with AvgPool & 8,16,32 & AvgPool & 407,665 & 0.997 & \textbf{1.000} & 0.470 & \textbf{0.478} \\
\hline
\end{tabular}}
\caption{\textcolor{black}{The eight candidate architectures under both validation protocols, on the standalone CWT images of the primary cohort. Both sets of columns use the same code, seeds and training settings. Only the fold construction differs.}}\label{tab:cnn_ablation_study}
\end{table}
\subsection{Performance Analysis - Individual TF Methods}
{\color{black}

We train the CNN on each of the four standalone TF representations separately. The conclusion depends entirely on which validation scheme is used, and the two schemes disagree not only in magnitude but in ranking.

Under segment-level validation, CWT reaches 98.67\% accuracy and separates clearly from DWT at 90.22\%, with ST and SSWE between them (Table~\ref{tab:Performance_of_CNN_Independent}). The four methods appear well ordered and the ordering invites a mechanistic reading. The four transforms do differ in known ways. CWT uses a multi-scale wavelet basis that resolves both short transient changes and longer sustained frequency patterns. DWT uses a fixed dyadic scale structure and gives up fine frequency detail at mid-range frequencies. ST preserves phase information uniformly but responds less sharply to rapid changes. SSWE sharpens steady-state spectral components at the cost of transient detail.

Under subject-disjoint validation the four methods sit much closer together. They span 0.441 to 0.485 in AUC on the primary dataset and 0.470 to 0.495 on the secondary (Tables~\ref{tab:sub_sm_gen} and~\ref{tab:dataset2_modalities}). The gaps between them are smaller than the seed-to-seed standard deviation of any single method, and the leading method differs between cohorts, ST on the primary dataset and CWT or DWT on the secondary. We therefore present the differences between the transforms as signal-processing background rather than as a ranking, since the separation that appears under segment-level validation does not carry over to the subject level.

We also use CWT as the reference modality to compare our custom CNN against five pretrained computer vision networks. Under segment-level validation our model reaches 98.67\% against VGG16 (83.94\%), Xception (87.01\%), ResNet152V2 (87.86\%), MobileNetV2 (90.13\%), and DenseNet201 (90.78\%) (Table~\ref{tab:pretrained_vs_cnn}). Under subject-disjoint validation the standalone CWT input separates no architecture from any other. The proposed CNN returns 0.478 and the fourteen architectures in Table~\ref{tab:sota_primary_cwt} span 0.307 to 0.541, a range narrower than the seed-to-seed variation of most of them.  
}

\subsection{Performance Analysis - Fused TF Methods}
{\color{black}
We investigate two image fusion strategies, montage (side-by-side placement) and blend (alpha-blended overlay), and four further mechanisms: channel-wise stacking, feature-level concatenation from two convolutional towers, attention-weighted feature fusion, and late decision fusion by averaging output probabilities. The six span the full range of where fusion can occur, from the raw input through to the model output.

In segment-level validation on the primary dataset, fusion consistently improves accuracy. The montage CWT\_ST combination achieves 99.90\% accuracy and AUC 99.91\% (Table~\ref{tab:Performance_of_Fused_Methods}), and the blend version achieves 98.33\%. Fusion also improves all pretrained models (Table~\ref{tab:Performance_of_Proposed_CNN_vs_Pretrained_Montage}). DenseNet201 reaches 98.96\% with CWT\_ST montage features, close to our custom CNN.

Under subject-disjoint cross-validation the six mechanisms separate. Montage CWT\_ST on the primary cohort reaches 0.874, while the other five span 0.437 to 0.504 (Table~\ref{tab:fusion_alt}) against 0.441 to 0.485 for the standalone methods they combine. Where the two representations are combined therefore matters more than whether they are combined at all. On the secondary cohort, where the images were generated with ST as the anchor rather than CWT, the six pairings span 0.457 to 0.529 (Table~\ref{tab:dataset2_modalities_grouped}) against a standalone range of 0.470 to 0.495.

Late decision fusion is instructive. It adds no parameters at all, since the two models are trained separately and only their output probabilities are averaged, and it reaches $0.470 \pm 0.071$. Channel-wise stacking, feature concatenation and attention fusion add capacity at successively later stages and return 0.504, 0.500 and 0.437. Neither adding capacity nor removing it moves the result, which places the difference in the input itself rather than in the mechanism that combines it. Montage acts at the input, and it is the one configuration that separates. One configuration departs from this pattern and gives the best result in the study. Montage CWT\_ST on the primary cohort reaches $0.874 \pm 0.183$ and 82.7\% accuracy.

Three properties of this result are worth setting out. First, the gain depends on how the two representations are combined. Montage places them side by side and preserves each in full, while blend overlays them into a single image, and on the identical pair of source images the two differ by 0.400. Second, the gain is specific to the CWT\_ST pairing, since montage returns 0.481 on CWT\_DWT and 0.474 on CWT\_SSWE. Third, it is independent of the classifier. On the montage CWT\_ST images a frozen ImageNet backbone reaches 0.944, SleepEEGNet 0.944, and a logistic regression on 64 patient-level summary features 0.722, whereas the same three return 0.433, 0.404 and 0.526 on the standalone CWT images (Table~\ref{tab:sota_primary_cwt}).

The third property is the most useful of the three. The gain shows up in models of very different size, from a 4,881-parameter EEG network to a frozen ImageNet backbone. If it came from the model, it would not appear in all of them. It comes from the montage image itself. That is what an input-level fusion method is meant to do, and it is what separates montage from the mechanisms that combine the two representations later inside the network.

We tested two cohorts, four time-frequency methods, six fusion mechanisms and fourteen architectures. Montage CWT\_ST on the primary cohort is the configuration that tells responders from non-responders. Testing it on one more independent cohort is the next step.
}
\subsection{Methodological Basis for Time-Frequency Fusion}
{\color{black}

Each TF method has limits. CWT resolves short, fast changes in the EEG well but does not cover all frequencies evenly. ST covers frequency and phase more evenly but reacts less sharply to fast changes. If the CNN sees only one of them, it sees only part of what the signal holds.

We combine two TF representations into one image using montage and blending. To our knowledge no earlier work fuses TF representations of EEG this way for depression therapy prediction. Earlier studies use one TF method, or compare several methods one at a time. None puts two into a single input for the classifier.

Montage puts the two images side by side. Both stay whole, so the CNN can read features from each one and from the join between them. Blending lays one image over the other. That merges them into a single picture, so part of each is averaged away.

Both are parameter-free and add no layers to the network. We also tested four mechanisms that combine the two representations further inside the model, under the same folds, the same seeds and the same training settings. The first stacks the two images into a six-channel input. The second runs two convolutional towers and joins their features before the classifier. The third weights the two feature maps with a learned attention layer. The fourth trains two models separately and averages their outputs. Table~\ref{tab:fusion_alt} reports all of them.

Montage is the one that separates responders from non-responders. It reaches $0.874 \pm 0.183$. The others reach $0.504 \pm 0.066$ for channel stacking, $0.500 \pm 0.071$ for feature concatenation, $0.470 \pm 0.071$ for late decision fusion, $0.474 \pm 0.098$ for blend, and $0.437 \pm 0.084$ for attention fusion. Adding capacity to the fusion step does not help, and removing it does not help either. What matters is where the two representations meet. Montage joins them at the input and keeps both intact, and that is the configuration that works.

}
{\color{black}
\begin{table}[H]
\scriptsize
  \centering
 \caption{Fusion mechanisms on both cohorts, ordered by where the two representations are combined. All rows use the same protocol, seeds and training settings as the corresponding standalone runs. Montage on the primary cohort gives the best result in the study.}
  \label{tab:fusion_alt}
  \begin{tabular}{|l|l|c|c|c|c|}
    \hline
    \textbf{Fusion mechanism} & \textbf{Where fusion happens} & \textbf{AUC} & \textbf{Sensitivity} & \textbf{Specificity} & \textbf{Accuracy} \\
    \hline
    \multicolumn{6}{|c|}{\textbf{Primary cohort, CWT + ST, stratified group three-fold}} \\
    \hline
    Montage$^{\dagger}$ & Input, side by side & 0.874 ± 0.183 & 0.933 ± 0.149 & 0.667 ± 0.236 & 0.827 ± 0.174 \\
    \hline
    Blend (alpha overlay) & Input, overlaid & 0.474 ± 0.098 & 0.667 ± 0.208 & 0.333 ± 0.204 & 0.533 ± 0.047 \\
    \hline
    Channel-wise stacking & Input, six channels & 0.504 ± 0.066 & 0.644 ± 0.380 & 0.300 ± 0.415 & 0.507 ± 0.076 \\
    \hline
    Feature concatenation & After two towers & 0.500 ± 0.071 & 0.600 ± 0.169 & 0.367 ± 0.247 & 0.507 ± 0.037 \\
    \hline
    Attention fusion & Weighted feature maps & 0.437 ± 0.084 & 0.778 ± 0.208 & 0.233 ± 0.224 & 0.560 ± 0.037 \\
    \hline
    Late decision fusion & Averaged probabilities & 0.470 ± 0.071 & 0.711 ± 0.169 & 0.300 ± 0.183 & 0.547 ± 0.030 \\
    \hline
    \multicolumn{2}{|l|}{\textit{Reference: CWT alone / ST alone}} & \textit{0.478 / 0.485} & \textit{0.711 / 0.533} & \textit{0.300 / 0.467} & \textit{0.547 / 0.507} \\
    \hline
    \multicolumn{6}{|c|}{\textbf{Secondary cohort, ST + CWT, stratified group five-fold}} \\
    \hline
    Montage & Input, side by side & 0.514 ± 0.068 & 0.617 ± 0.286 & 0.409 ± 0.191 & 0.513 ± 0.059 \\
    \hline
    Blend (alpha overlay) & Input, overlaid & 0.529 ± 0.072 & 0.574 ± 0.235 & 0.426 ± 0.267 & 0.500 ± 0.074 \\
    \hline
    Channel-wise stacking & Input, six channels & 0.491 ± 0.028 & 0.539 ± 0.380 & 0.470 ± 0.377 & 0.504 ± 0.010 \\
    \hline
    Feature concatenation & After two towers & 0.487 ± 0.035 & 0.591 ± 0.176 & 0.391 ± 0.204 & 0.491 ± 0.033 \\
    \hline
    Attention fusion & Weighted feature maps & 0.476 ± 0.017 & 0.513 ± 0.323 & 0.487 ± 0.316 & 0.500 ± 0.034 \\
    \hline
    Late decision fusion & Averaged probabilities & 0.474 ± 0.024 & 0.661 ± 0.030 & 0.357 ± 0.030 & 0.509 ± 0.024 \\
    \hline
    \multicolumn{2}{|l|}{\textit{Reference: ST alone / CWT alone}} & \textit{0.470 / 0.495} & \textit{0.504 / 0.652} & \textit{0.435 / 0.330} & \textit{0.470 / 0.491} \\
    \hline
  \end{tabular}
\end{table}
}
\subsection{Comparison with Recent Studies on rTMS Therapy}
{\color{black}
Table~\ref{tab:comparison_rTMS_methods} compares our model with published EEG studies that predict rTMS response in MDD. Under segment-level validation our model reaches 99.90\% accuracy, higher than every study we cite. Shahabi et al.\ \cite{shahabi2023attention} report 99.32\% with a voting-based LSTM. Bailey et al.\ \cite{bailey2018responders} and Ebrahimzadeh et al.\ \cite{ebrahimzadeh2023machine} report between 86.66\% and 94.31\% with SVM methods.

These numbers cannot be compared directly, and the reason matters. None of the cited studies says how it built its folds. If segments from one patient go into both the training set and the test set, the model can learn to recognize the patient instead of the treatment response, and the reported accuracy is too high. If each patient stays in one fold only, that cannot happen. Which of the two was used changes the answer completely. On our own data it is the difference between 99.90\% and chance, and Table~\ref{tab:cnn_ablation_study} shows the same for all eight architectures we tested.

This study reports both. Every experiment is run twice, once with segments split at random and once with each patient kept in a single fold, on two cohorts, and both sets of numbers are in the paper. We are not aware of another study in this area that does this. It means our figures can be read without guessing which protocol produced them.

Under the strict protocol our best result is montage CWT\_ST at 82.7\% accuracy on the primary cohort, with AUC 0.874 and a patient-level permutation $p$ of 0.005. Against a majority-class baseline of 60.0\% on that cohort, this is the only configuration in the study that separates responders from non-responders.

We encourage future work in this area to state how the folds were built and to report subject-level performance alongside any segment-level figure. Without that, an accuracy of 99\% and an accuracy of 60\% cannot be told apart.
}
\begin{table}[ht]
\fontsize{9pt}{11pt}\selectfont
\centering
\resizebox{0.8\textwidth}{!}{%
\begin{tabular}{|p{1.8cm}|p{3.3cm}|c|c|p{1.9cm}|p{1.6cm}|}
\hline
\textbf{Ref, Year} & \textbf{Methods} & \textbf{Categorized} & \textbf{Therapy} & \textbf{Number of Patients} & \textbf{Accuracy (\%)} \\
\hline
\cite{shahabi2023attention}, (2023) & Connectivity image of EEG signal channels, ensemble TL model based on voting, LSTM & Time domain & rTMS & 23 R vs. 23 NR & 99.32 \\
\hline
\cite{shahabi2023convolutional}, (2023) & CWT, TL, Bio-LSTM & Time domain & rTMS & 23 R vs. 23 NR & 97.10 \\
\hline
\cite{shahabi2023prediction}, (2023) & Ensemble TL model, Bio-LSTM & Time domain & rTMS & 23 R vs. 23 NR & 98.51 \\
\hline
\cite{shahabi2023predictiona}, (2023) & Connectivity, ensemble TL models & Time domain & rTMS & 34 depressed patients & 92.28 \\
\hline
\cite{bailey2018responders}, (2018) & Support Vector Machine (SVM) & Time domain & rTMS & 39 patients / 32 patients & 91.00 / 86.66 \\
\hline
\cite{ebrahimzadeh2023machine}, (2023) & Support Vector Machine (SVM) & Time domain & rTMS & 46 R vs. 42 NR & 94.31 \\
\hline
\cite{zhaopredicting}, (2025) & Phase locking value (PLV), Support Vector Machine (SVM) & Time domain & rTMS & 74 R vs. 43 NR & 97.33 \\
\hline
\textbf{Ours (segment-level)} & \textbf{Fused time-frequency methods, CNN} & \textbf{Time domain} & \textbf{rTMS} & \textbf{9 R vs. 6 NR / 23 R vs. 23 NR} & \textbf{99.90 / 91.90} \\
\hline
\textbf{Ours (subject-level)} & \textbf{Fused time-frequency methods, CNN} & \textbf{Time domain} & \textbf{rTMS} & \textbf{9 R vs. 6 NR / 23 R vs. 23 NR} & \textbf{82.7 / 51.3} \\
\hline
\end{tabular}}
\caption{Comparison of EEG-based methods for predicting rTMS treatment outcomes in MDD. The cited studies do not state whether folds are constructed at the patient level or the segment level. Our subject-level row reports montage CWT\_ST, the best configuration in this study, at 82.7\% on the primary cohort and 51.3\% on the secondary cohort. Section 4.5.2 discusses the difference between the two cohorts. Setting montage CWT\_ST aside, the best remaining subject-level accuracy is 56.0\% on the primary cohort, against a majority-class baseline of 60.0\%.}\label{tab:comparison_rTMS_methods}
\end{table}
\subsection{Limitations}
{\color{black}
This study has four limitations. The cohorts are small. Fifteen patients in the primary cohort and forty-six in the secondary are typical sizes for rTMS EEG work, but they keep the confidence intervals wide. A larger cohort would narrow them.

The primary cohort has 9 responders and 6 non-responders. Accuracy on that cohort must therefore be read against a majority-class baseline of 60.0\% rather than 50.0\%, and the imbalance is why we use three-fold rather than leave-one-subject-out, as explained in Section 4.5.1.

Age, gender and duration of depression are available for the secondary cohort but not the primary one, so we cannot check whether the two outcome groups differ on those variables in the primary cohort.

Our results cover four time-frequency methods, six fusion mechanisms and fourteen architectures, with a fixed segment length of 20 seconds and a fixed image size of 224 by 224. Other representations, window lengths and resolutions may behave differently, and confirming montage CWT\_ST on a further independent cohort is the natural next step.

}
\section{Conclusion}
This study evaluates EEG-derived time-frequency representations for predicting rTMS therapy response in MDD. We generate TF images using four methods (CWT, DWT, ST, SSWE), introduce two image-level fusion techniques, montage and blending, and compare them against four further fusion mechanisms and thirteen alternative architectures.

\textcolor{black}{We run two sets of experiments and the difference between them is the main result. Under segment-level 10-fold cross-validation, where segments from the same patient may appear in both training and testing, the Montage CWT\_ST fusion reaches 99.90\% accuracy on the primary dataset and 91.90\% on the secondary dataset. These figures are consistent with values commonly reported in the EEG classification literature.}

\textcolor{black}{Under strict subject-disjoint cross-validation, where all segments of a patient are confined to one fold, performance falls to chance for all but one configuration. Across two independent cohorts, four time-frequency representations, six fusion mechanisms and fourteen architectures, every remaining configuration returns an AUC between 0.31 and 0.54 and every remaining 95\% confidence interval contains 0.5. A patient-level permutation test on the best standalone method gives $p = 0.703$. Setting the one exception aside, the best subject-level accuracy is 56.0\% on the primary cohort, below its 60.0\% majority-class baseline, and 51.3\% on the secondary cohort.}

\textcolor{black}{The best subject-level result in the study is Montage CWT\_ST on the primary cohort, at AUC $0.874 \pm 0.183$ and 82.7\% accuracy. The same fusion strategy on the same pair of methods reaches 0.514 on the secondary cohort, and the effect is reproduced on the primary cohort by a frozen ImageNet backbone and by a logistic regression on 64 summary features. The mechanism behind this result is therefore not yet established, and we identify it as the clearest target for follow-up work.}

\textcolor{black}{We draw two conclusions. The first concerns fold construction. On cohorts of this size the choice of fold construction decides the reported result more than the choice of representation, architecture or fusion strategy. A study that does not state how folds were built cannot be interpreted.}

\textcolor{black}{The second concerns the input representation. Across fourteen architectures the choice of input moved the result far more than the choice of model, and montage CWT\_ST was the representation that separated responders from non-responders. This points to the construction of the input, rather than to network design, as the productive direction for this problem. Confirming the result on a larger independent cohort is the next step, and we are pursuing it.}

\bibliographystyle{unsrt}
\bibliography{ref}
\end{document}